\documentclass{article}

\usepackage{arxiv}

\usepackage[T1]{fontenc}
\usepackage{authblk}
\usepackage{graphicx}
\usepackage{epsfig,graphics,subfig}
\usepackage{color}
\usepackage{amsmath}
\usepackage{amssymb}
\usepackage{bbding}
\usepackage{amsbsy}
\usepackage{multirow}
\usepackage{tabularx}
\usepackage{mathtools}
\usepackage{array, booktabs}
\usepackage{graphicx}

\usepackage{algorithm}
\usepackage{algorithmicx}
\usepackage{algpseudocode}

\usepackage{appendix}

\newtheorem{lemma}{Lemma}

\usepackage{hyperref}
\usepackage{url}
\usepackage{cite}
\usepackage[numbers,sort&compress]{natbib}
\title{Towards Robust Neural Networks\\ via Orthogonal Diversity}
\author[1]{Kun Fang}
\author[2]{Qinghua Tao}
\author[1]{Yingwen Wu}
\author[1]{Tao Li}
\author[3]{Jia Cai}
\author[3]{Feipeng Cai}
\author[1]{Xiaolin Huang}
\author[1]{Jie Yang}
\affil[1]{Department of Automation, Shanghai Jiao Tong University\\
{\tt\small
\{fanghenshao,yingwen\_wu,li.tao,xiaolinhuang,jieyang\}@sjtu.edu.cn}}

\affil[2]{ESAT-STADIUS, KU Leuven, Belgium\\
{\tt\small qinghua.tao@esat.kuleuven.be}}

\affil[3]{Central Media Technology Institute, Huawei Technologies Ltd.\\
{\tt\small \{caijia1,caifeipeng\}@huawei.com}}

\date{}

\begin{document}

\maketitle

\begin{abstract}
Deep Neural Networks (DNNs) are vulnerable to invisible perturbations on the images generated by adversarial attacks, which raises researches on the adversarial robustness of  DNNs.
A series of methods represented by the \textit{adversarial training} and its variants have proven as one of the most effective techniques in enhancing the DNN robustness.
Generally, adversarial training focuses on enriching the training data by involving perturbed data.
Such data augmentation effect of the involved perturbed data in  adversarial training does not contribute to the robustness of  DNN itself and usually suffers from clean accuracy drop.
Towards the robustness of DNN itself, we in this paper propose a novel defense that aims at augmenting the model in order to learn features that are adaptive to diverse inputs, including adversarial examples.
More specifically, to augment the model, multiple paths are embedded into the network, and an orthogonality constraint is imposed on these paths to guarantee the diversity among them.
A margin-maximization loss is then designed to further boost such DIversity via Orthogonality (DIO).
In this way, the proposed DIO augments the model and enhances the robustness of DNN itself as the learned features can be corrected by these mutually-orthogonal paths.
Extensive empirical results on various data sets, structures and attacks verify the stronger adversarial robustness of the proposed DIO utilizing model augmentation. Besides, DIO can also be flexibly combined with different data augmentation techniques ({\it e.g.}, TRADES and DDPM), further promoting robustness gains.
\end{abstract}


\section{Introduction}\label{sec:introduction}


In recent decades, DNNs have been widely studied in various fields and  achieved record-breaking performance on numerous tasks \citep{goodfellow2016deep}.
However, the  robustness of DNNs also raised great concerns.
For example, on the classification task in computer vision, an image added with carefully-designed and visually-imperceptible perturbations can  fool  well-trained DNNs, {\it i.e.}, the \emph{victim}, to give incorrect predictions \citep{goodfellow2014explaining}.
Such perturbed images are known as \textit{adversarial examples}, and the process of generating adversarial examples is called \textit{adversarial attacks}. 
There have been various adversarial attacks, which can be mainly categorized into the types of  \textit{white-box} attacks \citep{goodfellow2014explaining,madry2018towards} and \textit{black-box} attacks \citep{papernot2017practical,liu2017delving}.
White-box attackers have access to the complete information of the victim, including the network architecture, parameters, gradients, etc. While in black-box attacks, only the input and the output of the victim  are available to the attackers.

Nowadays, it has been widely acknowledged that the most efficient way to improve the adversarial robustness shows to be the  \textit{adversarial training} \citep{goodfellow2014explaining,madry2018towards,qian2022survey}. 
Adversarial examples are generated during the training phase and at the same time they are also immediately utilized as the training data, so that DNN can learn knowledge of potential adversarial attacks.
In recent years, there have been many variants of adversarial training serving with state-of-the-art performances, in which   more effective ways are proposed to generate adversarial examples \citep{zhang2019theoretically,zhang2020geometry,yu2023improving}.

\begin{figure*}[!t]
\centering
\subfloat[Comparison of the architectures of a regular network (top) and the proposed DIO with $L$ heads (bottom).]{\includegraphics[width=0.31\linewidth]{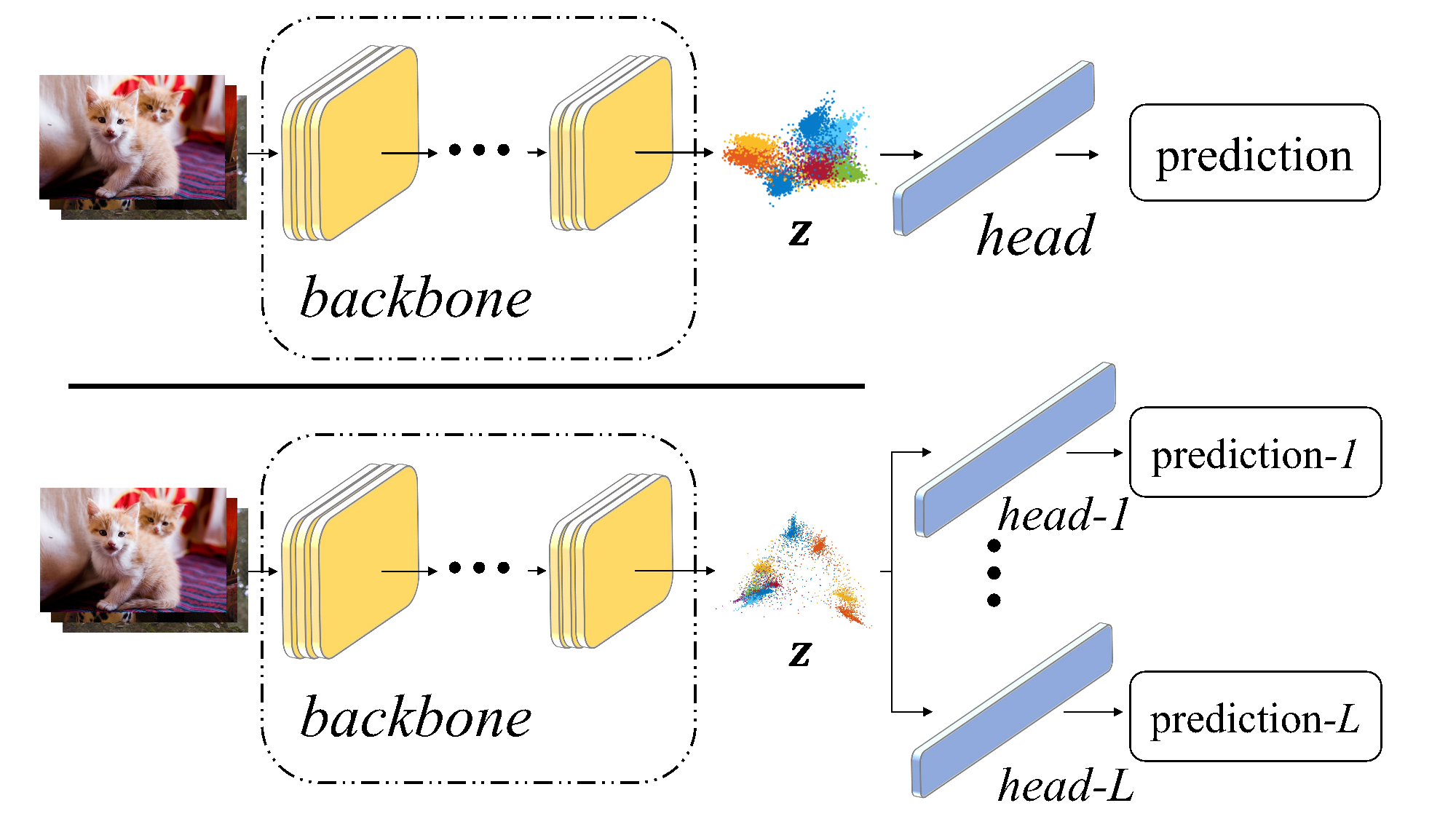}%
\label{fig:arch-comparison}}
\hfil
\subfloat[Learned features of a regular network and the DIO at the 1st convolution layer on the clean and perturbed images.]{\includegraphics[width=0.28\linewidth]{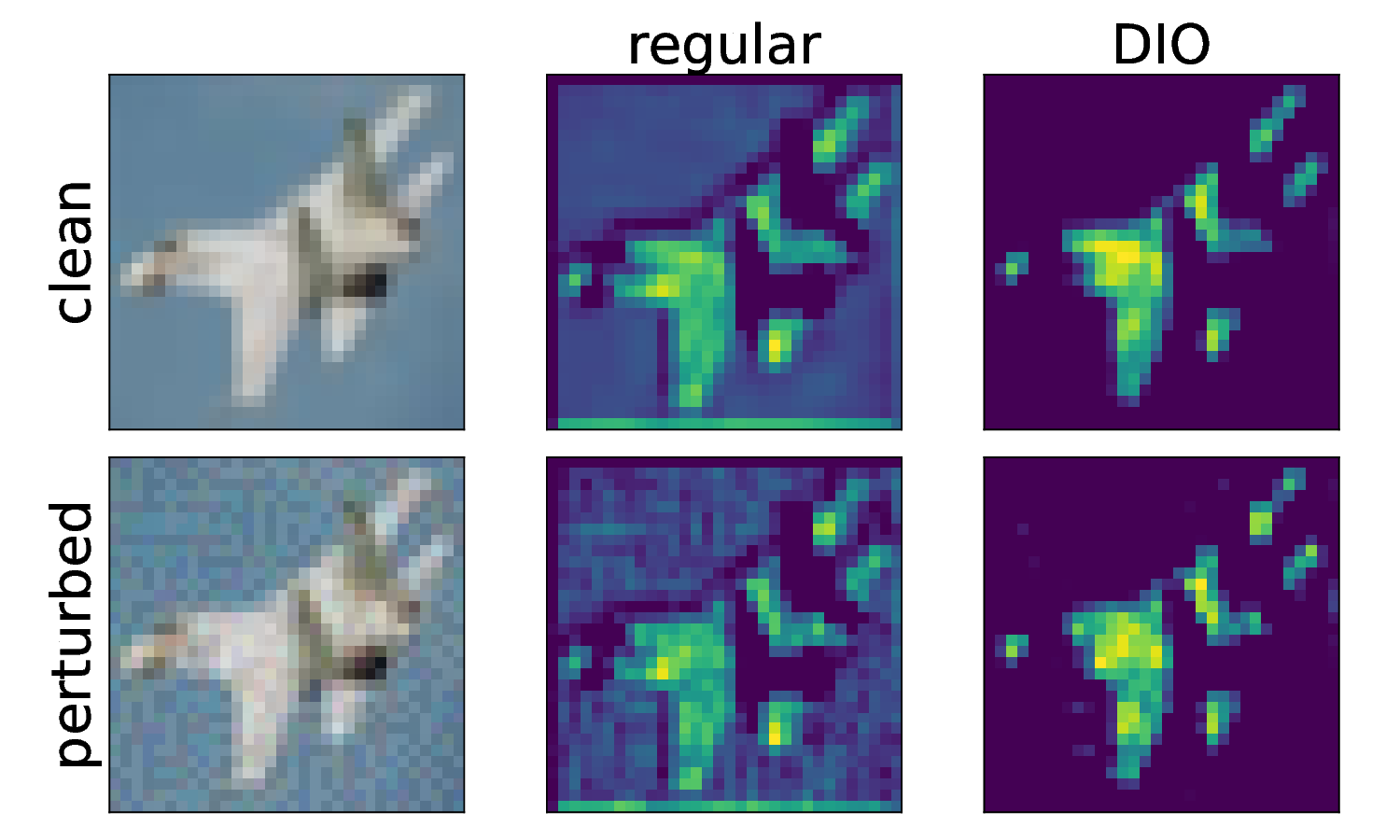}%
\label{fig:feature-first}}
\hfil
\subfloat[Learned features of a regular network and the DIO by the backbone on the clean and attacked images.]{\includegraphics[width=0.3\linewidth]{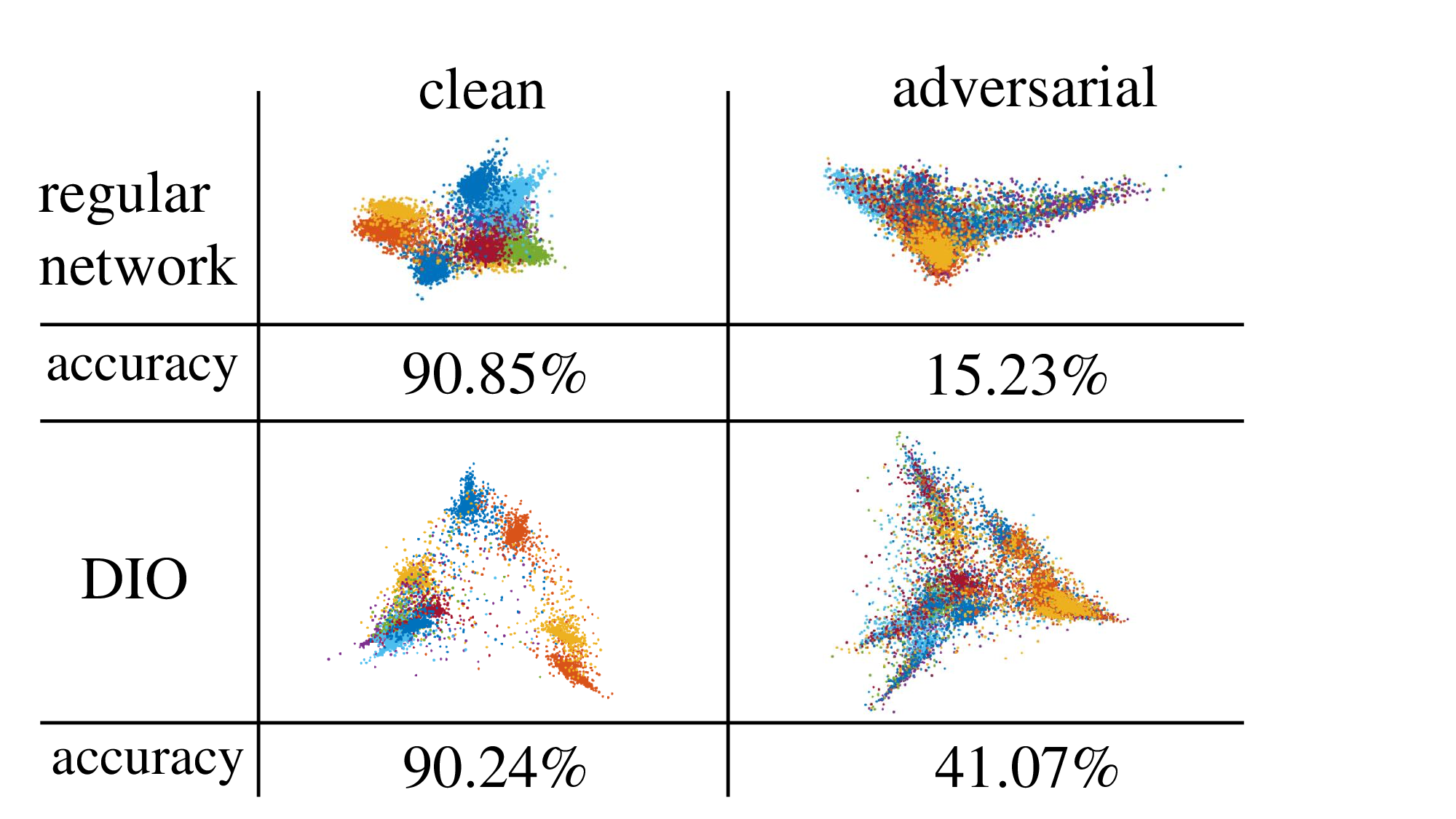}%
\label{fig:feature-last}}
\hfil
\caption{Comparisons of the architectures and the learned features between a regular network and the proposed DIO.}
\label{fig:intro}
\end{figure*}

Although adversarial training has shown effectiveness in defending adversarial attacks, the improved robustness intrinsically comes from the enriched data and thus it can be regarded as a \textit{data augmentation} method. 
However, the improved robustness comes from the trade-off between clean examples and adversarial ones, since the incorporation of adversarial examples also results in poor generalization performances on clean data and  weak abilities in defending various black-box attacks \citep{chen2020more,xing2021generalization}.
Therefore, it is of great interest to investigate another critical aspect influencing the  robustness, {\it i.e.}, the robustness of model itself, which is significant in understanding the essence and mechanism of adversarial robustness.
Hence, it can be much more meaningful to shed light on this valuable and yet ignored aspect, {\it i.e.}, the  defenses  exploiting the in-depth properties of the model itself to improve robustness, rather than simply incorporating data augmentation in training.

In this paper, we  address the robustness of DNNs from a novel aspect, {\it i.e.}, \textit{model augmentation}.
For the architecture, we propose to embed multiple paths into a regular DNN, where the  shared network layers and the augmented paths are trained to be adaptive to each other.
In this way, the augmented paths are simultaneously trained to learn diversified features fed to the subsequent network layers which are required to sustain consistently good performance on these features.
In the meantime, the shared  layers before the augmented paths are optimized to extract  features which can well adapt to  the augmented multiple paths, rather than a single path in regular DNNs.
Similar to data augmentation, the key point of our network augmentation is to endow these paths with diversity, meaning that such paths should differ to each other, though they share the common goal of attaining good predictions. 
Thus, in the proposed method, the augmented paths are orthogonal to each other, {\it i.e.}, the parameters of each paired paths are required to be  \textit{orthogonal}. Moreover, to enhance the significance of such orthogonality in high dimensions, a margin-maximization loss is designed by introducing distance constraints, which  collaborates with the orthogonality constraints across paths and jointly contributes to more effective diversity for robustness. Accordingly,
this proposed defense method is named as \textbf{DIO}: \textbf{DI}versity via \textbf{O}rthogonality, considering its working mechanism.

For DNNs, it commonly holds no prior knowledge on input data, that is, it makes no difference in processing  clean and adversarial samples, so that the informativeness of  features learned in the network is also unknown. 
To enhance the adaptiveness towards the unknown input images, the solution of DIO is to \textit{correct} the learned features via  constraints on the parameters.
Specifically, DIO augments orthogonal paths, so that the network layers can be trained  to \textit{correct} the learned features  through   mutual collaborations of multiple paths and their resulted diversity, as supported by the backward correction \citep{allen2023backward}, showing  that the training in higher-level layers improves the feature learning in lower-level layers through the  backward propagation in optimization. 
In this sense, on the one hand,  all these orthogonal paths get simultaneously optimized to promote more informative features learned in lower-level layers, where similar features of  clean and adversarial samples are extracted to avoid cumulative and amplified perturbations propagated across layers. 
On the other hand, the extracted features can well adapt to the augmented paths all achieving  accurate predictions, where notably more separable and diversified features of clean and perturbed samples can be learned, eventually contributing to  adversarial robustness improvements.

For instance, given a regular network and the corresponding DIO model both trained using clean samples only (see Fig.\ref{fig:arch-comparison} for an example of architecture comparison), we visualize the learned features at the first convolution layer: In Fig.\ref{fig:feature-first}, when a clean image (1st row) gets adversarially perturbed (2nd row, generated by attacking a third neural network), the features learned by a regular network change a lot (2nd column), while the learned features by DIO show much weaker perturbations (3rd column).
Besides, the features right before the multiple paths (the features $\mathbf{z}$ in Fig.\ref{fig:arch-comparison}) are also visualized in Fig.\ref{fig:feature-last}.
A white-box attack on the regular network messes up these features, resulting in a nearly random-guess accuracy.
In contrast, for the DIO model, these features still manifest certain separability against the same white-box attack.
Such feature variations at different positions indicate the robustness improvements, which does not come from the data augmentation effect, but the orthogonal diversity inside the model.
In this way, the augmented diversity on the network architectures corrects the learned features and also  enhances their adaptiveness towards both clean and adversarial samples via mutually-orthogonal paths, thereby boosting the adversarial robustness.

In numerical experiments, extensive comparisons are conducted to present solid evaluations on the effectiveness and  superior robustness improvements of the proposed DIO addressing network augmentation. 
In the cases only containing  clean samples in training, DIO substantially outperforms those non-data-augmented defenses. 
Equipped with those data-augmented techniques in training ({\it e.g.}, TRADES \citep{zhang2019theoretically} and DDPM \citep{ho2020denoising}), DIO also maintains higher robust accuracy.
Besides, we conduct 2 adaptive attacks and ablation studies to further verify the adversarial robustness of the proposed DIO.

The contributions of this work are summarized below:
\begin{itemize}
    \item We propose an adversarial defense  from a novel perspective of \textit{augmenting the model} to increase the adaptiveness to diverse inputs, which is essentially different from  data-augmented defenses.
    \item To achieve such adaptiveness, we embed multiple paths into the network and require  these paths to be \textit{mutually-orthogonal} for diversity inside the network.
    An associated margin-maximization loss is further designed to collaborate with the orthogonality constraint for more effective diversity.
    \item Extensive empirical results against white-box and black-box attacks on various network structures and data sets indicate the superior robustness of DIO. Adaptive attacks and ablation studies further investigate the properties and indispensable significance of the diversity via model augmentation.
\end{itemize}

The rest of this paper is organized as follows. Sec.\ref{sec:related-work} outlines related works on the adversarial attack and defense. Sec.\ref{sec:dio} describes details of the proposed DIO defense, including its framework, optimization, training and inference. Sec.\ref{sec:exp} shows the comprehensive empirical results. 
Concluding remarks and discussions are presented in Sec.\ref{sec:conclusion}.

\section{Related work}
\label{sec:related-work}

\subsection{Adversarial attack}
The existing adversarial attacks can be divided into  white-box and black-box attacks, which differ in the availability to the model information. 
The white-box attacks can be further categorized into two types,  gradient-based attacks and optimization-based ones.
The former exploits the parameter gradients to directly generate perturbed images, {\it e.g.}, Fast Gradient Sign Method (FGSM, \citep{goodfellow2014explaining}), Basic Iterative Method (BIM, \citep{kurakin2016adversarial}), Iterative  Least-Likely Class (ILLC, \citep{kurakin2016adversarial}), and Projected Gradient Descent (PGD, \citep{madry2018towards}).
The latter resolves optimization problems to obtain adversarial examples, such as C\&W attack \citep{carlini2017towards} and DeepFool \citep{moosavi2016deepfool}.

Given a clean sample and its label $(\mathbf{x},y)$, a DNN $f(\cdot)$ and a loss function $\mathcal{L}(\cdot,\cdot)$, the generations of adversarial examples $\mathbf{\hat{x}}$ of these attacks are briefly outlined as follows.

\noindent\textbf{FGSM} \citep{goodfellow2014explaining} moves a clean sample with the distance $\epsilon$ in the direction of the sign of the gradient:
\begin{equation}
    \label{eq:FGSM}
    \mathbf{\hat{x}} = \mathbf{x} + \epsilon\cdot\mathrm{sign}(\nabla_{\mathbf{x}}\mathcal{L}(f(\mathbf{x}),y)).
\end{equation}
BIM \citep{kurakin2016adversarial} and ILLC \citep{kurakin2016adversarial} are two iterative variants of FGSM. 

\noindent\textbf{PGD} \citep{madry2018towards} is considered as one of the strongest attack by far.
Starting from clean images added with random uniform noise, {\it i.e.}, $\mathbf{\hat{x}}^{(0)}=\mathbf{x}+\mathcal{U}(0,\epsilon)$, the $(t+1)$-th iteration of PGD is
\begin{equation}
    \label{eq:PGD}
    \mathbf{\hat{x}}^{(t+1)} = P_\epsilon\Big\{\mathbf{\hat{x}}^{(t)} + \gamma\cdot\mathrm{sign}(\nabla_{\mathbf{x}}\mathcal{L}(f(\mathbf{\hat{x}}^{(t)}),y))\Big\},
\end{equation}
where $P_\epsilon$ is the projection to the set $\{\mathbf{\hat{x}}|\|\mathbf{\hat{x}}-\mathbf{x}\|_\infty\leq\epsilon\}$ and $\gamma$ is the step size in each iteration.

\noindent\textbf{C\&W attack} \citep{carlini2017towards} solves the following optimization problem:
\begin{equation}
    \label{eq:cw-1}
    \mathop{\min}_\zeta\quad\Vert\frac{1}{2}(\tanh(\zeta)+1)-\mathbf{x}\Vert+c\cdot A(\frac{1}{2}(\tanh(\zeta)+1)),
\end{equation}
with $\frac{1}{2}(\tanh(\zeta)+1)-\mathbf{x}$ as the perturbation, $c$ as the chosen constant and $A(\cdot)$  defined as:
\begin{equation}
    \label{eq:cw-2}
    A(\mathbf{\hat{x}})=\max(Z(\mathbf{\hat{x}})_y-\mathop{\max}_{k\neq y}(Z(\mathbf{\hat{x}})_k), -\kappa),
\end{equation}
where $Z(\cdot)_k$ denotes the logit value corresponding to the $k$-th class and $\kappa$ denotes the confidence constant.

Black-box attacks can be divided into  transferability-based \citep{papernot2017practical,liu2017delving} and query-based \citep{chen2017zoo,andriushchenko2020square,NEURIPS2020_90599c8f,bai2023query}.
The former relies on the transferability of the adversarial examples, in which the adversarial examples are generated via a substitute model and then are transferred to fool the victim model.
The latter is designed to obtain the model output by iteratively querying the victim model.
The queried pairs of  input and output are then used either to train a surrogate model \citep{NEURIPS2020_90599c8f,bai2023query} or to directly estimate the gradients \citep{chen2017zoo,andriushchenko2020square} for generating the adversarial examples.

\subsection{Adversarial defense}
Adversarial training aims at improving the network robustness through  data augmentation by involving perturbed images into training data.
Currently PGD-based adversarial training is one of the most effective and popular  choices, achieving the highest accuracy under various attacks \citep{madry2018towards,athalye2018obfuscated}.
Other variants of adversarial training have also been proposed,  including but not limited to reduce the generalization gap \citep{chen2020more,xing2021generalization} or to alleviate the computation time of generating PGD-based adversarial examples \citep{zhang2019theoretically,zhang2020geometry}.
In addition, defenses  applying  pre-processings or transformations on the inputs can also be viewed as data augmentation techniques \citep{xie2018mitigating,guo2018countering}.

Aside from  data augmentation, another line of defenses aims at exploiting model properties 
by characterizing the network with specific constraints from various perspectives.
For example, several defenses introduce randomness into the network via adding unlearnable \citep{liu2018towards} or learnable noises on parameters \citep{liu2018adv,he2019parametric,wu2020adversarial} or activations \citep{jeddi2020learn2perturb} to improve robustness.
Others propose to take advantages of the ensemble of individual networks, images or weak defenses \citep{he2017adversarial,pang2019improving}.
Various constraints on the learned features have also been designed, {\it e.g.}, squeezing \citep{xu2017feature}, filtering \citep{xie2019feature}, or disentangling \citep{mustafa2019adversarial,cui2021learnable} the features.
There are also defenses that design denoising techniques on the input images to filter the perturbations \citep{liao2018defense,jia2019comdefend}.

\section{Diversity via orthogonality}
\label{sec:dio}

\textbf{Notations} A DNN $f(\cdot):\mathcal{R}^d\to \mathcal{R}^K$ can be decomposed into two parts: the backbone $g(\cdot):\mathcal{R}^d\to \mathcal{R}^m$ and the head $h(\cdot):\mathcal{R}^m\to \mathcal{R}^K$.
The backbone $g$ consists of cascaded convolution layers, and extracts features $\mathbf{z}\in\mathcal{R}^m$ from input images $\mathbf{x}\in\mathcal{R}^d$, {\it i.e.}, $\mathbf{z}=g(\mathbf{x})$.
Given these features, the head $h$ is a linear layer and outputs the predictive $K$-dimensional logits in correspondence to $K$ classes, {\it i.e.}, $h(\mathbf{z})=h(g(\mathbf{x}))=f(\mathbf{x})$.
Specifically, the head $h$ serves as a linear classifier with interconnection weights $\mathbf{W}\in\mathcal{R}^{m\times K}$ and bias $\mathbf{b}\in\mathcal{R}^K$, such that $h(\mathbf{z})=\mathbf{W}^T\mathbf{z}+\mathbf{b}$.

\subsection{Framework}
We first build the framework of DIO, and then formulate the optimization problem involved in the training.
The main idea of DIO is to improve the adaptiveness in handling diverse inputs via augmenting the model, which learns more informative and adaptive features, so that it can be more viable  to recognize  adversarial examples with different perturbations.

To augment the model, structurally, multiple paths are embedded after the backbone of a regular DNN, and then the network is trained to adapt to these augmented paths, boosting diversity inside the network. 
In each path, the backbone is connected to a subsequent  linear layer before the network outputs,  resulting in an augmented multi-head framework of DIO.
Thus, multiple heads $h^i$ are embedded in parallel and share the same backbone $g$.
Each augmented head is optimized to correctly  classify the features $\mathbf{z}$, which are learned by the backbone and are  jointly affected by the augmented multiple heads.
In this way, each head $h^i$  collaborates with the front backbone $g$ to form a regular DNN $h^i(g(\cdot))$.
An example of such a multi-head architecture is illustrated in Fig.\ref{fig:arch-comparison}.
After such network augmentation, another critical issue is how to endow diversity on the multiple paths.
Otherwise these paths can be optimized to converge to formulate the same.
The diversity among the paths is then achieved by posing well-designed constraints on the parameters of these heads, which are introduced in detail in the next subsection.

\begin{figure}[!t]
\centering
\includegraphics[width=0.5\linewidth]{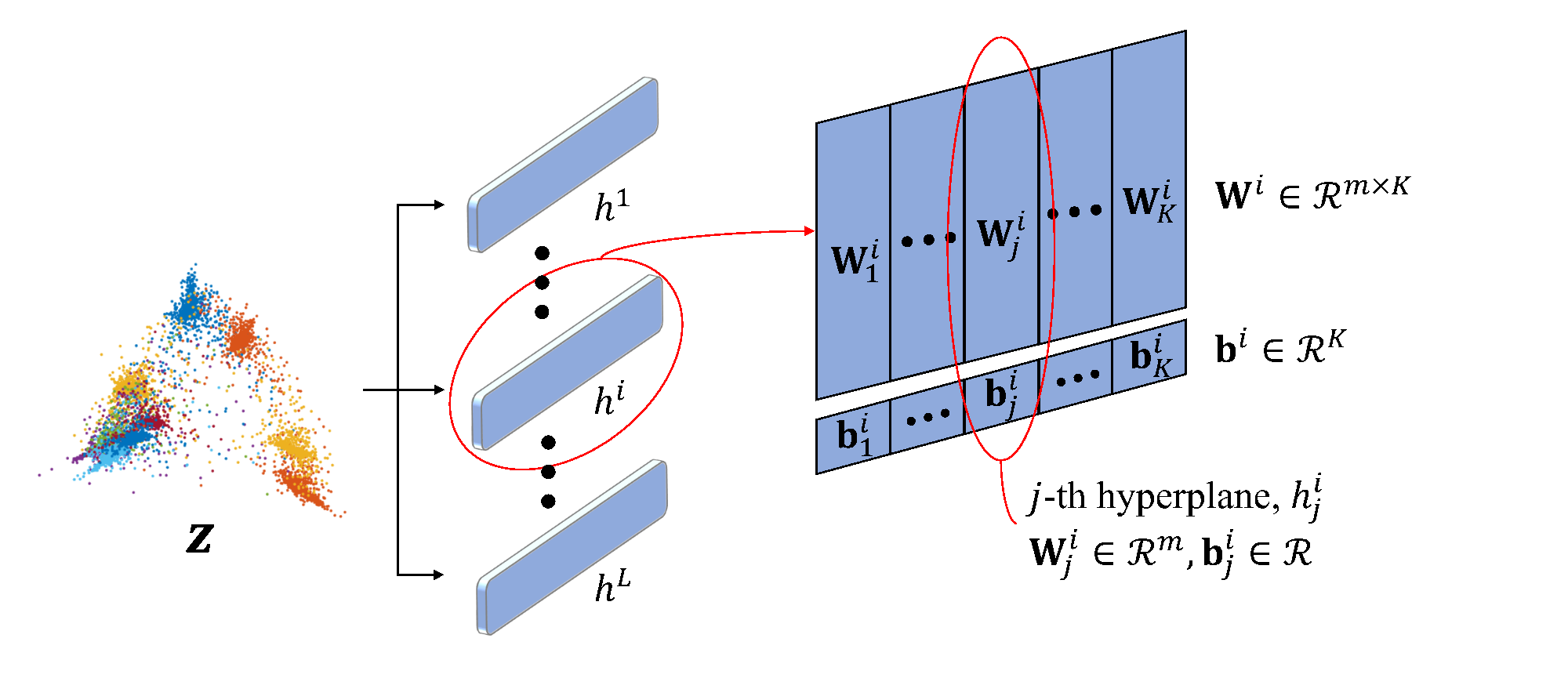}
\caption{Details of the multiple heads of DIO. The backbone is omitted for simplicity.}
\label{fig:arch-head}
\end{figure}

Before stepping into the optimization objective, more notations are introduced.
The  $i$-th head in DIO is denoted as $h^i$, with the associated parameters $\mathbf{W}^i\in\mathcal{R}^{m\times K}$ and $\mathbf{b}^i\in\mathcal{R}^K$.
The $j$-th hyperplane of  $h^i$  is denoted as $h^i_j$  {\it w.r.t.} the $j$-th class in the head $h^i$, with the associated parameters $\mathbf{W}_{j}^i\in\mathcal{R}^m$ and $\mathbf{b}_j^i\in\mathcal{R}$, where $j=1,2,...,K$.
In other words, $\mathbf{W}_j^i$ is the $j$-th column in $\mathbf{W}^i$, and $\mathbf{b}_j^i$ is the $j$-th element in $\mathbf{b}^i$.
Note that $h_y$ denotes the hyperplane {\it w.r.t.} the ground-truth class.
Fig.\ref{fig:arch-head} illustrates the details of the multiple heads of DIO and the related notations.

\subsection{Optimization}

On the basis of the multi-head structure, we now endow diversity among heads with two well-designed constraints.
Firstly, an orthogonality constraint is imposed on these heads, resulting in \textit{mutually-orthogonal} heads.
This constraint guarantees the inner diversity within the whole network and also outer diversity across  augmented heads.
Secondly, a distance constraint is imposed to counteract the nature that two random high-dimensional vectors are easy to approach orthogonal.
This distance constraint collaborates with the orthogonality constraint to achieve better diversity and robustness.
In short, the two constraints posed on the heads inject diversity into the network, which backward corrects the learned features in front layers (backbone), so as to attain greater robustness.
In the remainder of this section, we elucidate the optimization objective of DIO step by step.

\subsubsection{Orthogonality constraint}
Generally, two  vectors are orthogonal if their inner product is zero.
For two matrices or tensors, we can vectorize them and then perform the operation of inner product. 
Similarly, for two linear classifiers $h^i$ and $h^j$, their inner product is computed as:
\begin{equation}
    \label{eq:inner-prod}
    \langle h^i, h^j \rangle \equiv \langle\mathrm{vec}(\mathbf{W}^i),\mathrm{vec}(\mathbf{W}^j)\rangle.
\end{equation}
Therefore, two heads are orthogonal if the inner product of their vectorized weight matrices equals to zero.
Notice that the bias parameters in the heads are omitted in the computation here.

With the orthogonality constraint on augmented heads, we have the following optimization problem:
\begin{equation}
    \label{op:dio-ortho-1}
    \begin{array}{rl}
    \mathop{\min}\limits_{g,h^1,h^2,...,h^L} & \quad \sum_{i=1}^L\mathcal{L}_{ce}(h^i(g(\mathbf{x})), y)\\
    \mathrm{s.t.} &\quad\langle h^i,h^j\rangle=0,\quad\forall i\neq j,\quad i,j=1,2,...,L.
    \end{array}
\end{equation}
In Eq.\eqref{op:dio-ortho-1}, $L$ denotes the number of heads, and $\mathcal{L}_{ce}(\cdot,\cdot)$ denotes the cross entropy loss function.
The objective function in Eq.\eqref{op:dio-ortho-1} indicates that the sum of the classification losses of all  heads and the backbone is minimized, which   guarantees the classification performance of all heads.
The equality constraints in Eq.\eqref{op:dio-ortho-1} require that these heads to be mutually-orthogonal for diversity.
Parameters to be optimized include the backbone $g$ and all the heads $h^i$.

By leveraging the equality constraints into the objective, we have the following unconstrained optimization problem:
\begin{equation}
    \label{op:dio-ortho-2}
    \mathop{\min}_{g,h^1,h^2,...,h^L}\mathcal{L}_{c}+\alpha\cdot\mathcal{L}_o,
\end{equation}
with 
\begin{equation*}
\left\{
\begin{array}{l}
      \mathcal{L}_{c}=\sum_{i=1}^L\mathcal{L}_{ce}(h^i(g(\mathbf{x})), y),\\
    \mathcal{L}_o=\sum_{i=1}^{L}\sum_{j=1,j\neq i}^{L}\langle h^i,h^j\rangle ^2,
\end{array}
\right.
\end{equation*}
where the coefficient $\alpha$  controls the penalty on the orthogonality loss $\mathcal{L}_o$.

There exists another line of researches \citep{bansal2018can,jia2019orthogonal} that restrict the DNN parameters to be (nearly) orthogonal, {\it i.e.}, forcing the Gram matrix of the whole parameter matrix to be close to identity.
The orthogonality in these works does not require a multi-path structure, but instead aims to stabilize the parameter distribution and facilitate the optimization, which intrinsically differs from ours.

\subsubsection{Distance constraint}

The intuitive idea of the orthogonality constraint is to restrict that the learned features $\mathbf{z}$ can be simultaneously well-separated by those mutually-orthogonal classifiers. However, for high dimensional vectors, {\it i.e.}, the case in DNNs, the orthogonality constraint is relatively weak, since the inner product of two random high-dimensional vectors is generally quite small. 
To counteract such effect, we further design a distance constraint to enhance the orthogonality constraint and the diversity among these augmented paths.

A linear classifier $h$ consists of $K$ hyperplanes $h_j,j=1,\cdots,K$, each with the weight $\mathbf{W}_j\in\mathcal{R}^m$ and bias $\mathbf{b}_j\in\mathcal{R}$, corresponding to the $K$ classes.
The distance between the features $\mathbf{z}\in\mathcal{R}^m$ extracted from the backbone $g$ and the $j$-th hyperplane $h_j$ in the head $h$ is  defined as follows:
\begin{equation}
    \label{eq:dist-z-hyperplane}
    d(\mathbf{z},h_j)\equiv \frac{|\mathbf{W}_{j}^T\mathbf{z}+\mathbf{b}_{j}|}{\|\mathbf{W}_{j}\|}, 
\end{equation}
Based on $d(\mathbf{z},h_j)$, the margin between features $\mathbf{z}$ and the head $h$ is further defined as:
\begin{equation}
    \label{eq:dist-z-head}
    \begin{aligned}
    D(\mathbf{z},h)
    &\equiv\mathop{\min}_{j\neq y}\frac{|(\mathbf{W}_{j}^T\mathbf{z}+\mathbf{b}_{j})-(\mathbf{W}_{y}^T\mathbf{z}+\mathbf{b}_{y})|}{\|\mathbf{W}_{j}\|},\\
    &=\mathop{\min}_{j\neq y}\frac{|h_j(\mathbf{z})-h_y(\mathbf{z})|}{\|\mathbf{W}_{j}\|}\\
    &=\mathop{\min}_{j\neq y}\frac{|h_j(g(\mathbf{x}))-h_y(g(\mathbf{x}))|}{\|\mathbf{W}_{j}\|},
    \end{aligned}
\end{equation}
where $\mathbf{W}_y$ and $\mathbf{b}_y$ denote the weight and bias of the hyperplane $h_y$ {\it w.r.t.} the ground-truth class.

The distances between $\mathbf{z}$ and each hyperplane $h_j$ in the head $h$ can be measured with $d(\mathbf{z},h_j)$.
Thus, by taking a normalized difference between $d(\mathbf{z},h_j)$ and $d(\mathbf{z},h_y)$, this metric $D(\mathbf{z},h)$ actually measures the required minimal distance to change the output of the head to a wrong prediction.

The complete optimization problem of DIO is constructed as follows:
\begin{equation}
    \label{op:dio-complete}
    \mathop{\min}_{g,h^1,h^2,...,h^L}\mathcal{L}_{c}+\alpha\cdot\mathcal{L}_o+\beta\cdot\mathcal{L}_d,
\end{equation}
with 
\begin{equation*}
\left\{
\begin{array}{l}
       \mathcal{L}_{c}=\sum_{i=1}^L\mathcal{L}_{ce}(h^i(g(\mathbf{x})), y),\\
    \mathcal{L}_o=\sum_{i=1}^{L}\sum_{j=1,j\neq i}^{L}\langle h^i,h^j\rangle ^2,\\
    \mathcal{L}_d=\dfrac{1}{L}\sum_{i=1}^L\max\left\{0,\tau-\mathop{\min}_{j\neq y}\frac{|h^i_j(g(\mathbf{\bar{x}}))-h^i_{y}(g(\mathbf{\bar{x}}))|}{\|\mathbf{W}^i_{j}\|}\right\}.
\end{array}
    \right.
\end{equation*}

In Eq.\eqref{op:dio-complete}, the margin-maximization loss $\mathcal{L}_d$ involving the margin term in Eq.\eqref{eq:dist-z-head} is taken with a similar form to the hinge loss.
For the backbone $g$ and a given head $h^i$, only those correctly-classified samples $(\mathbf{\bar{x}},y)$ are incorporated in calculating the loss $\mathcal{L}_d$, {\it i.e.}, $h^i(g(\mathbf{\bar{x}}))=y$.
The margin $D(\mathbf{\bar{z}},h)$ is then calculated based on the backbone $g$, the head $h^i$, and the specified samples $(\mathbf{\bar{x}},y)$, which measures the minimal distance required to make the current head output a false prediction.
For a robust network, this margin $D(\mathbf{\bar{z}},h)$ should be maximized as much as possible so that the perturbations accumulated from the input will not cause the network to respond incorrectly.
Therefore, we design such a form for the $\mathcal{L}_d$: the maximization in the hinge loss and the outermost minimization works together to form a min-max optimization problem, which forces the margin $D(\mathbf{\bar{z}},h)$ as close to the hyper-parameter $\tau$ as possible.
The coefficient $\beta$ balances the effect of $\mathcal{L}_d$.
In addition, a theoretical discussion on the relationships between the two constraints is provided in the {\it supplementary material}.

\subsection{Training and inference}

\begin{figure}[!t]
\centering
\includegraphics[width=0.9\linewidth]{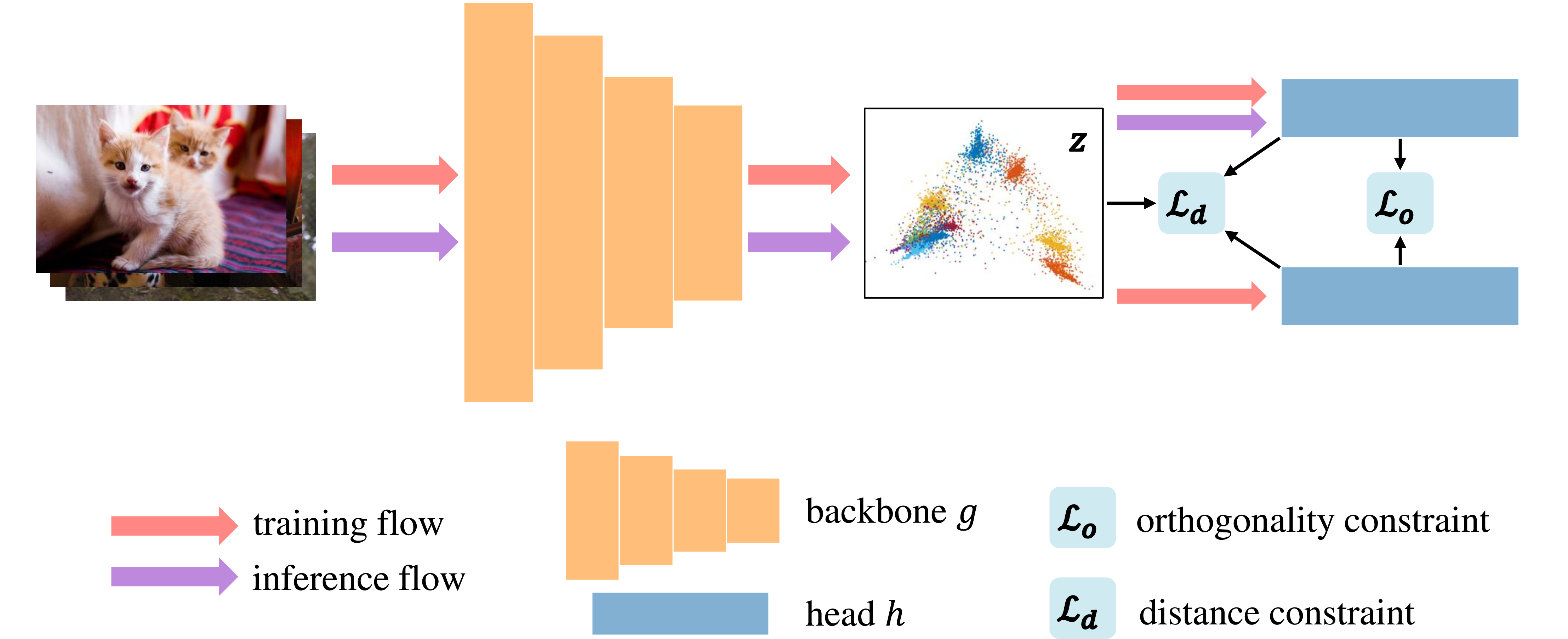}
\caption{An illustration on the training and inference of the proposed adversarial defense DIO with a 2-head network structure as an example.
During training, all the heads are involved into the forward propagation.
The orthogonality constraint $\mathcal{L}_o$ is calculated on the 2 heads, while the distance constraint $\mathcal{L}_d$ is determined based on the 2 heads and the learned features $\mathbf{z}$ from the backbone.
The cross-entropy loss is omitted here for simplicity.
In inference, only one head is randomly selected to give predictions.}
\label{fig:response-scheme}
\end{figure}

The training and inference flows of DIO are outlined in Fig.\ref{fig:response-scheme}, and details are presented in the following two subsections.

\subsubsection{Training}
In the training of DIO, the losses are computed and then all the parameters of the backbone and heads get updated by stochastic gradient descent.
Among the three losses $\mathcal{L}_c$, $\mathcal{L}_o$ and $\mathcal{L}_d$, the cross entropy loss $\mathcal{L}_c$ is determined by the input data, the backbone and the multiple heads.
The orthogonality loss $\mathcal{L}_o$ is calculated only on the $L$ heads.
The distance loss $\mathcal{L}_d$ is determined based on all the heads and the features of those correctly-classified samples.

The training of DIO only requires clean images, and DIO can also be flexibly equipped with other data-augmented defenses involving additional adversarial examples, together to achieve greater robustness (see Section \ref{sec:exp} for details).
An algorithm illustrating the training details of DIO can be found in the {\it supplementary material}.

\subsubsection{Inference}
\label{sec:inference}
When the training phase of DIO is finished, we acquire a DNN with $L$ mutually-orthogonal heads.
Each of these $L$ heads  collaborates with the shared backbone to form a regular DNN $f^i=h^i(g(\cdot)),i=1,2,...,L$.
The resulting $L$ networks exhibit almost the same performance in both the generalization on unseen test samples and the adversarial robustness on adversarial ones.
Numerically, these mutually-orthogonal heads hold nearly the same predictive accuracy on the input images (see Section \ref{sec:exp} for details), indicating  that the diversity augmentation equally benefits every head.

In the inference phase, during every forward process, a head is randomly chosen to give the prediction output.

\noindent\textbf{Remark}\quad In this section, the framework of the proposed adversarial defense DIO has been presented in detail, including the multi-head structure, novel losses, training and inference strategies.
DIO aims to exploit model properties to enhance adversarial robustness through orthogonality and distance constraints on the network parameters, so that it essentially differs from other model-based defenses, which commonly adopt the techniques of ensemble \citep{pang2019improving}, random noise \citep{he2019parametric} and feature modifications (such as squeezing \citep{xu2017feature}, filtering \citep{xie2019feature}, disentangling \citep{mustafa2019adversarial}).
We can note that DIO is not an ensemble-based defense as there is no ensemble strategy on the multiple heads {\it in inference}.
Besides, DIO also differs from data-based defenses that aim at generating adversarial examples for data augmentation during training \citep{zhang2019theoretically,zhang2020geometry};
for example, the AugMax \citep{wang2021augmax} unifies both hard and diverse data as a stronger augmentation technique, which is a completely different route from our model-based adversarial defense.

\section{Experiments}
\label{sec:exp}

\subsection{Setups}
\noindent\textbf{Data sets:} We evaluate the performance of the proposed DIO on CIFAR10 \citep{krizhevsky2009learning}, CIFAR100 \citep{krizhevsky2009learning}, and TinyImageNet \citep{deng2009imagenet}.
There are 50K training samples and 10K test samples of $32\times32\times3$ colored images categorized into 10 classes and 100 classes in CIFAR10 and CIFAR100, respectively.
TinyImageNet is a subset of ImageNet \citep{deng2009imagenet} with 100K training samples and 10K validation samples of 200 classes.

\noindent\textbf{Network architectures:} Multiple popular network architectures are used to  evaluate the effectiveness of the proposed  DIO, including VGG networks \citep{simonyan2014very}, pre-activation residual networks \citep{he2016identity}, and wide residual networks \citep{zagoruyko2016wide}.

\noindent\textbf{Adversarial attacks:} For white-box attacks, we consider two gradient-based, $l_\infty$-norm attacks, FGSM \citep{goodfellow2014explaining} and PGD \citep{madry2018towards} attacks, and one optimization-based, $l_2$-norm attack, C\&W attack \citep{carlini2017towards}.
For black-box attacks, we evaluate the model robustness against the transferability-based attack \citep{liu2017delving} and the query-based attack SQUARE \citep{andriushchenko2020square}.
Besides, we also evaluate the performance under an ensemble attack AutoAttack (AA, \citep{croce2020reliable}), which includes three white-box attacks (APGD-CE, APGD-DLR and FAB \citep{croce2020minimally}) and one black-box attack (SQUARE).
 
In the following, the terminology \textit{vanilla training}  refers to the  training only using clean samples, while the terminology  \textit{adversarial training} denotes the training also involving adversarial examples.
All source code is based on PyTorch \citep{paszke2019pytorch} and has been released publicly\footnote{\href{https://github.com/fanghenshaometeor/DIversity-via-Orthogonality}{https://github.com/fanghenshaometeor/DIversity-via-Orthogonality}}.

\subsection{Diversity benefits robustness}
\label{sec:diversity-benefits-robustness}

\begin{table*}[t]
	\centering
	\caption{Accuracy (\%) of \textbf{vanilla-trained}, \textbf{non-data-augmented} defenses of \textbf{PRN18} on clean and adversarial images generated via white-box (FGSM and C\&W) and black-box (TRANSFER and SQUARE) attacks from the test sets in {CIFAR10} and {CIFAR100}. The best performance is highlighted in {bold}.}\label{tab:robustness-comparison-1}
	\begin{tabular}{cc|c|cc|cc}
		\toprule
        data set & defense & clean & {\bf FGSM} & {\bf C\&W} & {\bf TRANSFER} & {\bf SQUARE} \\
		\midrule
		\multirow{7}{*}{CIFAR10} &
		baseline 
		& 94.64 & 25.31 & 29.77 & 25.31 & 15.93 \\
		& DeNoise-GA 
		& 94.31 & 27.15 & 28.62 & 38.39 & 14.30 \\
		& DeNoise-MA 
		& $\mathbf{94.78}$ & 24.25 & 27.42 & 37.77 & 14.13 \\
		& DeNoise-MD
		& 94.63 & 24.36 & 28.50 & 39.27 & 15.45 \\
		& PNI 
		& 94.43 & 27.73 & 34.57 & 37.42 & \textbf{60.42} \\
		& PCL
		& 83.66 & 30.06 & 40.93 & 30.89 & 24.58 \\
		& DIO 
		& 94.59$\pm$0.01 & {\bf38.57$\pm$0.33} & {\bf41.64$\pm$0.29} & {\bf39.92$\pm$0.93} & 45.33$\pm$0.22 \\
        \midrule
		\multirow{7}{*}{CIFAR100} &
		baseline 
		& \textbf{74.94} & 6.41  & 15.48 & 6.41  & 8.42\\
		& DeNoise-GA 
		& 74.52 & 5.92 & 15.16 & 17.04 & 8.10\\
		& DeNoise-MA 
		& 74.58 & 6.12 & 15.54 & 16.59 & 7.63\\
		& DeNoise-MD
		& 74.66 & 6.34 & 15.66 & 17.01 & 8.43\\
		& PNI 
		& 74.87 & 6.28 & 16.47 & 15.74 & 27.49\\
		& PCL
		& 56.50 & 11.16 & 17.12 & 17.02 & 6.05\\
		& DIO
		& 74.17$\pm$0.10 & {\bf19.13$\pm$0.17} & {\bf25.04$\pm$0.16} & {\bf19.79$\pm$0.06} & {\bf48.67$\pm$0.21}\\
		\bottomrule
	\end{tabular}
\end{table*}

In this part, empirical experiments are presented to verify the robustness improvements benefited from the augmented diversity brought by the proposed DIO.
On the one hand, we compare  our method  with other \textit{non-data-augmented} adversarial defenses under the setting of \textit{vanilla training}.
On the other hand, we demonstrate that our proposed DIO can cooperate with the SOTA \textit{data-augmented} and \textit{adversarial-trained} defenses to further boost robustness.
Besides, two adaptive attacks are deliberately designed to comprehensively evaluate the robustness of the proposed DIO.
In the following experiments, the standard deviations of DIO indicate the variances of the recognition results of the $L$ paths. The standard deviations of multiple runs for all the methods are omitted since they are very small ($<0.50\%$) and hardly affect the results.

\subsubsection{Comparisons on non-data-augmented defenses}

Several representative non-data-augmented defenses are selected in experiments, including PNI \citep{he2019parametric}, DeNoise \citep{xie2019feature} and PCL \citep{mustafa2019adversarial}. Here,
PNI \citep{he2019parametric}  regularizes the network by adding learnable Gaussian noise on the parameters, and can be viewed as a randomness-based methods.
DeNoise \citep{xie2019feature} is a denoising-based defense via filters for the perturbations on features.
PCL \citep{mustafa2019adversarial} is a feature-based defense which promotes features of the same class to be close and  features of difference classes to be far away to disentangle the feature proximity.

Table \ref{tab:robustness-comparison-1} illustrates the comparison results using the network architecture PreActResNet18 (PRN18) on CIFAR10 and CIFAR100.
All compared defenses are reproduced based on their released source code.
The suffixes ``-GA'', ``-MA'' and ``-MD'' denote the Gaussian, mean, and median filters in DeNoise \citep{xie2019feature}, respectively.
We set the batch size as 256, the initial learning rate as 0.1 with 10\% decay at the 75-th and the 90-th epoch, and run totally 100 epochs on the training set.
The SGD optimizer is adopted with 0.9 momentum and $5e^{-4}$ weight decay.
Moreover, the adversarial examples  are not involved during the training, so as to reveal whether these non-data-augmented defenses themselves are able to improve robustness or not.

In Table \ref{tab:robustness-comparison-1}, for FGSM, the maximum perturbation $\epsilon$ is set as $8/255$ of $l_\infty$ norm.
For C\&W, the $l_2$ norm is adopted with confidence $\kappa=0$, 0.01 learning rate, 9 binary search steps and 3 maximum iterations.
For the transfer-based black-box attack, adversarial examples generated by FGSM with $l_\infty$ bound $8/255$ on the baseline model are transferred to deceive other models.
For the query-based SQUARE, the $l_2$ norm is adopted with bound $0.5$.

The results in Table \ref{tab:robustness-comparison-1}  illustrate the superiority of DIO in robustness improvements over other non-data-augmented defenses.
Against the white-box attacks, DIO shows the highest robust accuracy while DeNoise and PNI only bring minor gains.
Against the black-box attacks, DIO could defend not only the transferred adversarial examples but also the iterative queries.
In this experiment, DIO is verified to  substantially outperform other non-data-augmented defenses in the absence of adversarial examples, where such superior robustness  comes from the model augmentation specified in DIO.

\begin{table*}[t]
	\centering
	\caption{Accuracy (\%) of \textbf{adversarial-trained}, \textbf{data-augmented} defenses of \textbf{PRN18} and \textbf{WRN34X10} on clean and adversarial images generated via white-box (PGD-20 and PGD-100), black-box (SQUARE) and ensemble (AutoAttack) attacks from the test set in {CIFAR10}. The best performance is highlighted in {bold}. Results of both the {BEST} and {LAST} models during the training phase are recorded.} \label{tab:robustness-comparison-2}
	\resizebox{\textwidth}{!}{
	\begin{tabular}{cc|c|cc|c|c|c|cc|c|c}
		\toprule
		\multirow{2}{*}{defense} & \multirow{2}{*}{model}
		& \multicolumn{5}{c|}{BEST}
		& \multicolumn{5}{c}{LAST} \\
		&& \multicolumn{1}{c}{clean} 
		& \textbf{PGD-20} & \multicolumn{1}{c}{\textbf{PGD-100}} 
		& \multicolumn{1}{c}{\textbf{SQUARE}} & \textbf{AA} 
		& \multicolumn{1}{c}{clean} 
		& \textbf{PGD-20} & \multicolumn{1}{c}{\textbf{PGD-100}}  
		& \multicolumn{1}{c}{\textbf{SQUARE}} & \textbf{AA} \\
		\midrule
		AT & \multirow{2}{*}{PRN18}
		& 83.48 & 50.70 & 50.33 & 72.42 & 47.34
		& 84.51 & 47.17 & 46.71 & 72.12 & 44.63 \\
		DIO+AT
        & & $82.75\pm0.06$ 
        & $\mathbf{51.01\pm0.05}$ 
        & $\mathbf{50.60\pm0.07}$ 
        & $\mathbf{81.85\pm0.08}$ 
        & $\mathbf{49.36\pm0.11}$
        & $84.31\pm0.04$ 
        & $\mathbf{49.26\pm0.08}$ 
        & $\mathbf{48.64\pm0.07}$
        & $\mathbf{82.74\pm0.04}$ 
        & $\mathbf{47.33\pm0.06}$ \\
        \cmidrule{2-12}
		AT & \multirow{2}{*}{WRN34X10}
		& 86.48 & 52.10 & 51.81 & 73.88 & 49.96
		& 86.65 & 47.32 & 46.81 & 71.37 & 45.88 \\
		DIO+AT & 
		& $86.20\pm0.03$ 
		& $\mathbf{53.80\pm0.07}$ 
		& $\mathbf{53.33\pm0.06}$ 
		& $\mathbf{83.00\pm0.07}$ 
		& $\mathbf{52.05\pm0.06}$
		& $86.44\pm0.01$ 
		& $\mathbf{53.23\pm0.04}$ 
		& $\mathbf{52.68\pm0.03}$ 
		& $\mathbf{81.52\pm0.07}$ 
		& $\mathbf{51.53\pm0.04}$ \\
		\cmidrule{1-12}
        TRADES & \multirow{2}{*}{PRN18}
        & 82.10 & 52.85 & 52.58 & 71.05 & 49.29
        & 82.15 & 52.24 & 52.04 & 71.14 & 48.92 \\
		DIO+TRADES & 
		& $\mathbf{82.61\pm0.07}$ 
		& $\mathbf{52.87\pm0.05}$ 
		& $52.57\pm0.06$ 
		& $\mathbf{81.53\pm0.09}$ 
		& $\mathbf{50.50\pm0.09}$
		& $\mathbf{83.07\pm0.07}$ 
		& $\mathbf{52.32\pm0.06}$ 
		& $51.95\pm0.07$ 
		& $\mathbf{82.09\pm0.08}$ 
		& $\mathbf{49.86\pm0.09}$ \\
        \cmidrule{2-12}
        TRADES & \multirow{2}{*}{WRN34X10}
        & 84.96 & 56.19 & 55.98 & 72.74 & 52.93 
        & 85.25 & 54.59 & 54.28 & 72.28 & 52.14 \\
		DIO+TRADES & 
		& $\mathbf{85.26\pm0.04}$
		& $56.04\pm0.03$ 
		& $55.71\pm0.05$ 
		& $\mathbf{82.81\pm0.07}$ 
		& $\mathbf{53.33\pm0.06}$
		& $84.95\pm0.05$ 
		& $\mathbf{54.83\pm0.04}$ 
		& $\mathbf{54.50\pm0.04}$ 
		& $\mathbf{82.37\pm0.07}$ 
		& $\mathbf{52.17\pm0.05}$ \\
		\cmidrule{1-12}
		AWP & \multirow{2}{*}{PRN18}
		& 82.49 & 54.38 & 54.10 & 72.32 & 50.07
		& 82.40 & 54.18 & 53.87 & 72.38 & 48.94 \\
		DIO+AWP & 
		& $81.87\pm0.07$ 
		& $\mathbf{54.41\pm0.07}$ 
		& $\mathbf{54.12\pm0.04}$ 
		& $\mathbf{81.02\pm0.08}$ 
		& $\mathbf{51.66\pm0.08}$
		& $\mathbf{82.65\pm0.06}$ 
		& $\mathbf{54.33\pm0.05}$ 
		& $\mathbf{53.97\pm0.06}$ 
		& $\mathbf{81.62\pm0.07}$ 
		& $\mathbf{51.41\pm0.06}$ \\
		\cmidrule{2-12}
		AWP & \multirow{2}{*}{WRN34X10}
		& 87.22 & 57.93 & 57.49 & 75.91 & 54.23
		& 87.31 & 57.24 & 56.81 & 76.18 & 53.75 \\
		DIO+AWP & 
		& $87.00\pm0.04$ 
		& $\mathbf{58.05\pm0.05}$ 
		& $\mathbf{57.62\pm0.05}$ 
		& $\mathbf{85.00\pm0.05}$ 
		& $\mathbf{55.56\pm0.08}$
		& $\mathbf{87.64\pm0.03}$ 
		& $57.00\pm0.04$ 
		& $56.48\pm0.03$ 
		& $\mathbf{85.39\pm0.05}$ 
		& $\mathbf{54.45\pm0.03}$ \\
		\cmidrule{1-12}
        LBGAT & \multirow{2}{*}{PRN18}
        & 83.23 & 51.08 & 50.76 & 72.83 & 48.06 
        & 84.30 & 50.61 & 50.06 & 73.04 & 47.92 \\
		DIO+LBGAT & 
		& $82.08\pm0.05$ 
		& $\mathbf{51.28\pm0.04}$ 
		& $\mathbf{50.94\pm0.05}$ 
		& $\mathbf{80.22\pm0.10}$ 
		& $\mathbf{48.99\pm0.06}$
        & $82.41\pm0.04$ 
        & $\mathbf{51.14\pm0.08}$ 
        & $\mathbf{50.80\pm0.04}$ 
        & $\mathbf{80.42\pm0.04}$ 
        & $\mathbf{49.03\pm0.06}$ \\
        \cmidrule{2-12}
        LBGAT & \multirow{2}{*}{WRN34X10}
        & 86.45 & 52.77 & 52.28 & 74.87 & 50.64
        & 87.08 & 48.97 & 48.32 & 74.04 & 47.22 \\
		DIO+LBGAT & 
		& $84.64\pm0.03$ 
		& $\mathbf{53.89\pm0.06}$ 
		& $\mathbf{53.50\pm0.06}$ 
		& $\mathbf{82.03\pm0.07}$
		& $\mathbf{51.90\pm0.05}$
		& $86.27\pm0.01$ 
		& $\mathbf{51.44\pm0.06}$ 
		& $\mathbf{50.93\pm0.05}$ 
		& $\mathbf{81.79\pm0.06}$
		& $\mathbf{49.86\pm0.03}$ \\
		\cmidrule{1-12}
		GAIRAT & \multirow{2}{*}{PRN18}
		& 78.09 & 59.39 & 59.46 & 64.51 & 36.73 
		& 80.96 & 58.20 & 58.19 & 64.39 & 30.55 \\
        DIO+GAIRAT & 
        & $\mathbf{79.76\pm0.46}$ 
        & $\mathbf{62.45\pm0.29}$ 
        & $\mathbf{62.18\pm0.30}$ 
        & $\mathbf{79.70\pm0.46}$ 
        & $\mathbf{45.37\pm0.62}$
        & $79.34\pm0.80$
        & $\mathbf{62.44\pm0.27}$ 
        & $\mathbf{62.16\pm0.28}$ 
        & $\mathbf{79.30\pm0.80}$ 
        & $\mathbf{45.67\pm0.73}$  \\
        \cmidrule{2-12}
        GAIRAT & \multirow{2}{*}{WRN34X10}
        & 82.73 & 61.06 & 61.10 & 68.77 & 40.41
        & 83.20 & 59.13 & 59.12 & 67.47 & 38.04 \\
		DIO+GAIRAT & 
		& $81.13\pm0.93$ 
		& $\mathbf{67.04\pm0.39}$ 
		& $\mathbf{66.84\pm0.38}$ 
		& $\mathbf{81.04\pm0.94}$ 
		& $\mathbf{49.86\pm1.41}$
		& $81.66\pm0.77$ 
		& $\mathbf{67.06\pm0.35}$ 
		& $\mathbf{66.86\pm0.85}$ 
		& $\mathbf{81.62\pm0.77}$ 
		& $\mathbf{50.36\pm0.86}$ \\
		\bottomrule
	\end{tabular}
	}
\end{table*}

\subsubsection{Comparisons on data-augmented defenses}

In this experiment, our model-augmented DIO is plugged into  the data-augmented defenses to evaluate whether DIO  could further boost the adversarially trained defenses.
The selected powerful defenses include PGD-based adversarial training (AT, \citep{madry2018towards}), TRADES \citep{zhang2019theoretically}, GAIRAT \citep{zhang2020geometry}, LBGAT \citep{cui2021learnable}, and AWP \citep{wu2020adversarial}.
AT \citep{madry2018towards} uses the PGD attack to generate adversarial examples;
TRADES \citep{zhang2019theoretically} proposes that the predictions of noisy data should approximate that of clean data;
GAIRAT \citep{zhang2020geometry} re-weights the adversarial examples according to their distances towards the classification boundary during training.
The above three methods focus on designing sophisticated ways of generating adversarial examples.
For the other two defenses, LBGAT \citep{cui2021learnable} and AWP \citep{wu2020adversarial}, the former adopts another model trained on clean data to guide the training of the robust model and the latter designs a double-perturbation on both the weights and the data.

Table \ref{tab:robustness-comparison-2} illustrates the comparison results using network architectures PreActResNet18 and WideResNet34X10 (WRN34X10) on CIFAR10.
All compared defenses are reproduced based on their released source code.
For a fair comparison, we follow the widely-used settings in \citep{zhang2019theoretically,cui2021learnable}. For AT, TRADES and LBGAT, we run  100 epochs on the training set and adopt the SGD optimizer with 0.9 momentum, $5e^{-4}$ weight decay and an initial learning rate 0.1, where the learning rate is  divided by 10 at the 75-th and the 90-th epoch.
For AWP \citep{wu2020adversarial} and GAIRAT \citep{zhang2020geometry}, the corresponding defenses are trained following the settings reported in their original papers.
In particular, for AWP and LBGAT, the generation of  adversarial examples is by PGD attack of $l_\infty$ norm with 10 iterations, step size $2/255$, and  bound $8/255$, {\it w.r.t.} the cross entropy loss only.
In Table \ref{tab:robustness-comparison-2}, two strong white-box attacks, PGD-20 and PGD-100, are included into comparisons, together with a black-box attack (SQUARE) and an ensemble attack (AA).
For  PGD-20 and PGD-100, the bound is set as $8/255$ of $l_\infty$ norm with 20 and 100 iterations, respectively.
For SQUARE attack, its $l_2$ norm version is adopted and bounded by 0.5.
For AA attack, its $l_\infty$ norm version is adopted with bound $8/255$.

As indicated by Table \ref{tab:robustness-comparison-2}, those data-augmented techniques (AT, TRADES and GAIRAT) can be flexibly integrated into our proposed model-augmented defense DIO.
The diversity from those mutually-orthogonal parameters in DIO is empirically verified to well adapt to  the diversity reflected by the adversarial examples. 
On the other hand, by endowing orthogonal diversity into the network, DIO could also benefit those adversarially trained defenses (AWP and LBGAT).
Thus, as shown, it enables to further boost the robustness, even against strong  attacks in both white-box and black-box settings.
For instance, DIO of WRN34X10 architecture equipped with the data-augmented defense GAIRAT  achieves over 65\% robust accuracy against the white-box PGD-20 and PGD-100 attacks. 
The enhancement on the robustness against the black-box SQUARE attack is substantial, {\it i.e.},  achieving nearly  10\% accuracy improvement.

Due to the page limitation, we defer more empirical results to the {\it supplementary material}, including robustness comparisons with more defenses on more complex data sets (CIFAR100 \citep{krizhevsky2009learning}, TinyImageNet \citep{deng2009imagenet} and DDPM-generated data \citep{ho2020denoising}), thorough empirical results on the adaptive attacks against DIO, additional analyses on the model capacities, convergence and sensitivity, and a list of the hyper-parameters used in all the experiments.

\subsection{Ablation study}
Apart from the multi-head topology in the proposed DIO, the well-designed losses also play critical roles in improving robustness.
In this part, ablation studies are conducted to separately evaluate the effects of the two losses $\mathcal{L}_o$ and $\mathcal{L}_d$.
The settings of ablation study are listed in Table \ref{tab:settings-ablation-study}. 
All the models are trained under the same network structure with a fixed number of heads for a rigorous evaluation on the individual effectiveness of $\mathcal{L}_o$ and $\mathcal{L}_d$.

\begin{figure}[!ht]
\centering
\includegraphics[width=0.80\linewidth]{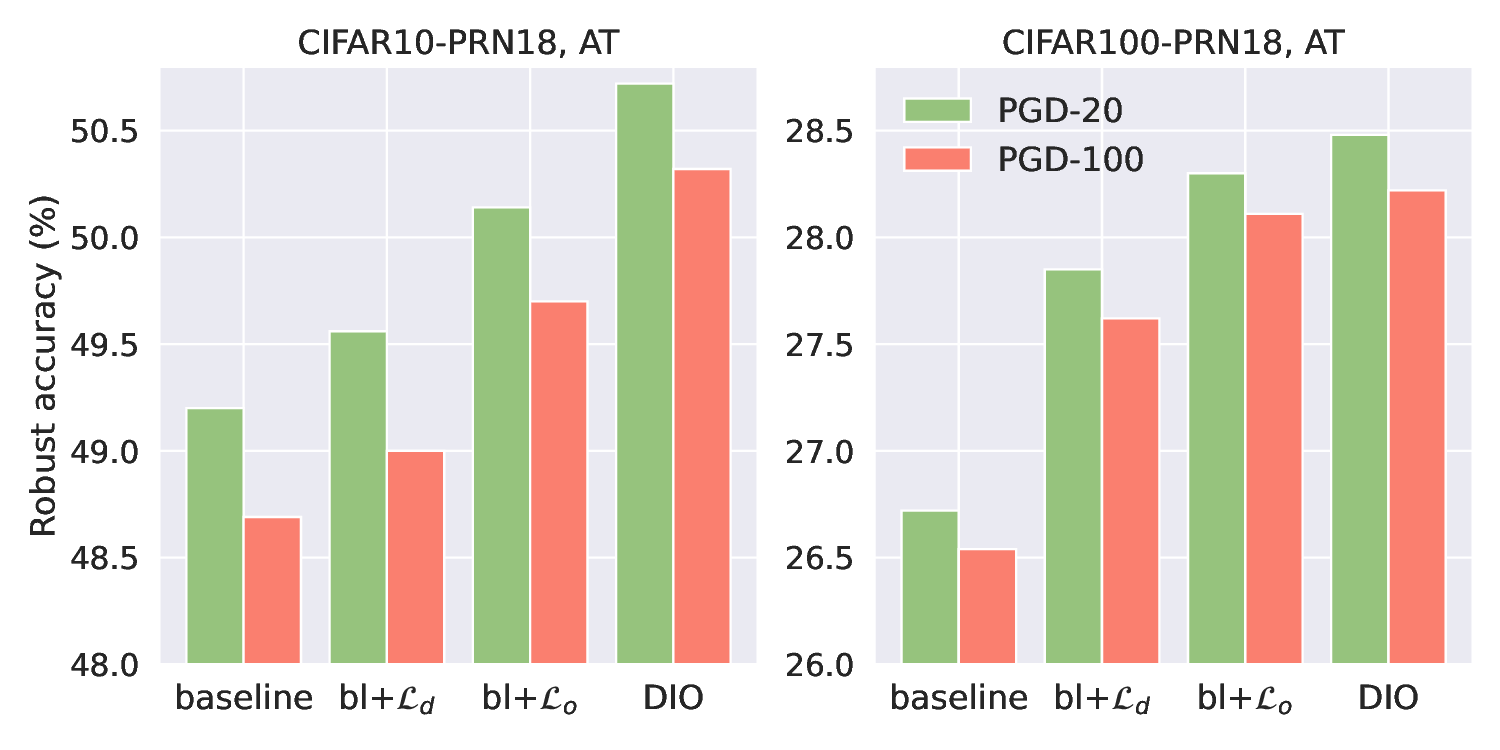}
\caption{Results of ablation studies. These models of PRN18 are adversarially trained on CIFAR10 and CIFAR100 and are evaluated against the PGD-20 and PGD-100 attacks.}
\label{fig:ablation-study}
\end{figure}

The ablation studies are executed on adversarially-trained models of PRN18 on CIFAR10 and CIFAR100.
Following the settings in Table \ref{tab:settings-ablation-study}, we train the multi-head models in correspondence to different losses and evaluate the model robustness against the PGD-20 and PGD-100 attacks, shown in Fig.\ref{fig:ablation-study}

\noindent\textbf{Individual effects of $\mathcal{L}_o$ and $\mathcal{L}_d$}\quad
In Fig.\ref{fig:ablation-study}, each of the two constraints can boost the robust accuracy on the baseline multi-head models.
$\mathcal{L}_o$ imposes orthogonality diversities on the head parameters so that the backbone could learn robust features.
$\mathcal{L}_d$ forces the learned features far away from the $L$ classifiers in the multi-head model, leading to stronger robustness.

\noindent\textbf{Joint effects of $\mathcal{L}_o$ and $\mathcal{L}_d$}\quad
The joint collaboration of $\mathcal{L}_o$ and $\mathcal{L}_d$ brings more robustness improvements on the multi-head model, shown by the highest robust accuracy in Fig.\ref{fig:ablation-study}.
The two constraints together contribute to more robust features that are far away from the $L$ mutually-orthogonal classifiers and further enhance the adversarial robustness of DIO models.

\begin{table}[!t]
	\centering
	\caption{Setups in  ablation studies. All the models are trained under the same network structure with a fixed number of heads. The ``\checkmark'' and ``{\tiny{\XSolidBrush}}'' indicate that the corresponding loss is included in training the multi-head models or not, respectively.}\label{tab:settings-ablation-study}
	\begin{tabular}{c|cccc}
		\toprule
		model & \# of heads & $\mathcal{L}_{ce}$ & $\mathcal{L}_o$ & $\mathcal{L}_d$ \\
		\midrule
		baseline & $L=10$ & \checkmark & {\scriptsize{\XSolidBrush}} & {\scriptsize{\XSolidBrush}} \\
		baseline+$\mathcal{L}_d$ & $L=10$ & \checkmark & {\scriptsize{\XSolidBrush}} & \checkmark \\
		baseline+$\mathcal{L}_o$ & $L=10$ & \checkmark & \checkmark & {\scriptsize{\XSolidBrush}} \\
		DIO & $L=10$ & \checkmark & \checkmark & \checkmark \\
		\bottomrule
	\end{tabular}
\end{table}

\section{Conclusion and limitations}
\label{sec:conclusion}
In this paper, we propose a novel adversarial defense algorithm.
Different from the common adversarially trained defenses relying on data augmentation from  adversarial examples, the proposed defense DIO aims at augmenting the model diversity to enhance the network robustness itself.
Structurally, multiple heads are designed within the network, and    share the same backbone.
Parametrically, the augmented heads are required to be mutually-orthogonal to achieve  model diversity,
and an associated distance constraint is established to alleviate the natural nearly orthogonality of two high-dimensional vectors, making the orthogonality constraint more meaningful.
Extensive empirical results verify the  robustness improvements of DIO in both the vanilla training and the adversarial training cases.
On the basis of powerful data augmentation techniques, \textit{e.g.}, DDPM, DIO could also achieve more robustness gains.
Adaptive attacks and ablation studies are further given to investigate the properties and  the significance of the model diversity of DIO in improving adversarial robustness from different aspects.
We hope this work could promote inspirations in investigating the model properties to improve robustness of the network itself, aside of designing sophisticated data augmentation techniques.

Meanwhile, there also exist several limitations of DIO. 
This model augmentation mechanism requires structural modifications, \textit{i.e.}, the augmented multiple heads, bringing additional costs in both training time and storage, though very minor.
Besides, such a multi-head structure limits its usage, as a standard network structure would be preferred in various downstream tasks, such as detection and segmentation.
Therefore, it would be interesting to explore a more efficient way to better characterize the orthogonality diversity in neural networks without a multi-head network structure.

\section*{Acknowledgments}

This work is jointly supported by the National Natural Science Foundation of China under Grants 62376153 and 62376155, and the Research Program of Shanghai Municipal Science and Technology Committee under Grants 22511105600.
Part of the computations in this paper were run on the Siyuan Mark cluster supported by the Center for High Performance Computing at Shanghai Jiao Tong University.

\bibliographystyle{unsrtnat}  
\bibliography{reference}

%








\clearpage

\begin{appendices}

\section*{\centering{\large Supplementary Material for \\ {Towards Robust Neural Networks via Orthogonal Diversity}} \\~}

\section{Algorithm}
\label{sec:alg}
The training algorithm of the proposed DIO is shown below.

\begin{algorithm}[h]
\caption{The training algorithm of the proposed adversarial defense DIO.}
\label{alg}
\begin{algorithmic}[1]
\Require Training set $X_{tr}=\{(\mathbf{x}_i,y_i)\}^n_{i=1}$,  number of heads $L$,  hyper-parameters $\alpha$, $\beta$ and $\tau$,  learning rate $\eta$,  number of epochs \textit{epoch}, flag of performing adversarial training \textit{adv-train}. Note that $\theta$ denotes the to-be-optimized parameters of $g,h^1,...,h^L$.
\Ensure A DNN with augmented heads $h^{i}(g(\cdot)),i\in\{1,2,...,L\}$.
\While {not reach \textit{epoch}}
\For {$(\mathbf{x},y)$ in $X_{tr}$}
\If{\textit{adv-train}}
\State Generate adversarial examples $(\mathbf{x}^{a},y)$
\State $loss\leftarrow\mathbf{compute\_loss}(h^i(g(\cdot)),(\mathbf{x}^{a},y),\tau)$
\Else
\State $loss\leftarrow\mathbf{compute\_loss}(h^i(g(\cdot)),(\mathbf{x},y),\tau)$
\EndIf 
\State $\theta\leftarrow\theta-\eta\cdot\nabla_\theta loss$\qquad$\triangleright$ \textit{Update parameters}
\EndFor
\EndWhile 
\State
\Function{compute\_loss}{$h^i(g(\cdot)),(\mathbf{x},y),\tau$}
\State Calculate $\mathcal{L}_{c}$ and $\mathcal{L}_{o}$; 
\State Initialize $\mathcal{L}_{d}=0$;
\For {each head $h^i$}
\State Find the correctly-classified $(\mathbf{\bar{x}},y)$;
\State Calculate $D(g(\mathbf{\bar{x}}),h^i)$;
\State Calculate $\mathcal{L}_{d}\gets\mathcal{L}_{d}+\frac{1}{L}\max\big\{0,\tau-D(g(\mathbf{\bar{x}}),h^i)\big\}$;
\EndFor
\State \Return $\mathcal{L}_{c}+\alpha\cdot\mathcal{L}_{o}+\beta\cdot\mathcal{L}_{d}$
\EndFunction
\end{algorithmic}
\end{algorithm}

\section{Theoretical discussion}

From a theoretical perspective,  discussions on the proposed orthogonality and distance constraints are given as follows, illustrating their effectiveness. Specifically, we will show how the margin-maximization loss $\mathcal{L}_d$ works and how it cooperates with the orthogonal loss $\mathcal{L}_o$ to boost informative orthogonality in high-dimensional space, by the help of the Vapnik-Chervonenkis (VC) dimension \citep{Vapnik1998} in statistical learning theory and the orthogonality Lemma induced from the Johnson-Lindenstrauss Lemma \citep{johnson1984extensions,dasgupta1999elementary,kaban2015improved}.

Considering the bi-classification task and linear classifier $h(\mathbf{x})=\mathrm{sign}(\mathbf{w}^T\mathbf{x}+b)$, the $\mathcal{L}_d$ measures the margin between the sample points $\mathbf{x}$ and the linear classifier $h(\cdot)$, which degenerates to the support vector machine \citep{cortes1995support}.
For the general case of the hypothesis space $\mathcal{H}_1$ of all linear classifiers in $\mathcal{R}^d$, 
The VC dimension equals to $d+1$, {\it i.e.}, $VC(\mathcal{H}_1)=d+1$.
Once the margin $\delta$ is introduced on the linear classifier $h$, the hypothesis space becomes  $\mathcal{H}_2$\footnote{If $\mathcal{H}_2$ is the space of all linear classifiers in $\mathcal{R}^d$ that separate the training data with margin at least $\delta$, then $VC(\mathcal{H}_2)\leq\min(\lceil\frac{R^2}{\delta^2}\rceil,d)+1\leq VC(\mathcal{H}_1)$, where $R$ is the radius of the smallest sphere in $\mathcal{R}^d$ that contains the data.}, and then the VC dimension of this hypothesis space $VC(\mathcal{H}_2)$ is reduced \citep{Vapnik1998}.
Next, the orthogonality lemma based on the Johnson-Lindenstrauss Lemma \citep{johnson1984extensions,dasgupta1999elementary,kaban2015improved} is given as follows.
\begin{lemma}
\label{lemma:ortho}
Given two vectors $\mathbf{u},\mathbf{v}\in\mathcal{R}^d$  independently  sampled from $\mathcal{N}(0,1/d)$  and a constant  $\varepsilon\in(0,1)$, we have
$$P(|\langle \mathbf{u},\mathbf{v}\rangle|\geq\varepsilon)\leq4\exp{\Big(-\frac{\varepsilon^2d}{8}\Big)}.$$
\end{lemma}

Lemma \ref{lemma:ortho} theoretically states that the inner product of two vectors, the measure of the orthogonality,  randomly sampled in a high dimensional space can be very low. Thus,  such two   vectors are easy to be nearly orthogonal. In our proposed margin-maximization loss $\mathcal{L}_d$,  a minimal distance is imposed on the output of the head to maintain the correct prediction. Such a distance loss can introduce a margin to a linear classifier, {\it i.e.}, $\mathbf{\hat{u}}$ and $\mathbf{\hat{v}}$ are sampled from a new set $A=\{\mathbf{w}|y_i(\mathbf{w}^T\mathbf{x}_i+b)>\delta\}$, instead of the $\mathcal{N}(0,1/d)$ in Lemma \ref{lemma:ortho}. Thus, 
the set $A$ represents the linear classifiers where the distance constraint $\mathcal{L}_d$ is posed on.
Associated with the reduced VC dimension and Lemma \ref{lemma:ortho}, one can conclude that
\begin{equation}\label{eq:neq:distance}
P(|\langle\mathbf{\hat{u}},\mathbf{\hat{v}}\rangle|\leq\varepsilon)\geq P(|\langle \mathbf{u},\mathbf{v}\rangle|\leq\varepsilon).
\end{equation}
The inequality in (\ref{eq:neq:distance}) indicates that the distance constraint $\mathcal{L}_d$  results in weaker orthogonality, which counteracts the natural nearly orthogonality told by  Lemma \ref{lemma:ortho} and thereby makes the orthogonality constraint $\mathcal{L}_o$ much more meaningful. Therefore, the combination of $\mathcal{L}_d$ and $\mathcal{L}_o$ is effective and solid to introduce orthogonality across the augmented heads (high-dimensional vectors) and to boost the performance of each augmented head (classifier).

\begin{table*}[!ht]
	\centering
	\caption{Results of clean and robust accuracy (\%) of various defenses on CIFAR10, CIFAR100 and TinyImageNet. All the clean accuracy ($\mathrm{Acc}_c$) is evaluated on the clean test set. The robust accuracy ($\mathrm{Acc}_r$) on CIFAR10 and CIFAR100 is evaluated against the AA attack and the $\mathrm{Acc}_r$ on TinyImageNet is evaluated against the PGD-20 attack. 
	The superscript $^\dagger$ indicates that the corresponding results are from the original paper \citep{cui2021learnable}.} \label{tab:robustness-comparison-3}
	\begin{tabular}{cc|c|cc}
		\toprule
		data set & model & defense & $\mathrm{Acc}_c$ & $\mathrm{Acc}_r$ \\
		\midrule
		\multirow{5}{*}{CIFAR10} & \multirow{2}{*}{WRN34X10} 
		         & LBGAT+TRADES$^\dagger$ 
		         & 81.98 & 53.14 \\
		         & & \textbf{DIO}+TRADES   
		         & $\mathbf{85.26\pm0.04}$ & $\mathbf{53.33\pm0.06}$ \\
		\cmidrule{2-5}
		         & \multirow{3}{*}{WRN34X20} 
		         & \citet{rice2020overfitting}$^\dagger$ 
	             & 85.34 & 53.42 \\
		         & & LBGAT+TRADES$^\dagger$
		         & 83.61 & 54.45 \\
		         & & \textbf{DIO}+TRADES
		         & $\mathbf{85.94\pm0.05}$ & $\mathbf{54.47\pm0.08}$ \\
		\midrule
		\multirow{11}{*}{CIFAR100} & \multirow{7}{*}{WRN34X10}
		         & AT           
		         & 59.87 & 27.07 \\
		         & & \citet{sitawarin2021sat}$^\dagger$
		         & 62.82 & 24.57 \\
		         & & \citet{chen2022efficient}$^\dagger$
		         & 62.15 & 26.94 \\
		         & & TRADES$^\dagger$       
		         & 56.50 & 26.87 \\
		         & & LBGAT+TRADES$^\dagger$
		         & 60.43 & 29.34 \\
		         & & \textbf{DIO}+TRADES
		         & $60.21\pm0.09$ & $\mathbf{29.58\pm0.23}$ \\
		         & & \textbf{DIO}+AT       
		         & $55.16\pm0.14$ & $\mathbf{30.86\pm0.18}$ \\
	    \cmidrule{2-5}
	             & \multirow{3}{*}{WRN34X20} 
	             & AT 
	             & 61.28 & 27.58 \\
	             & & LBGAT+TRADES$^\dagger$
	             & 62.55 & 30.20 \\
	             & & \textbf{DIO}+AT
	             & $57.81\pm0.15$ & $\mathbf{31.40\pm0.18}$ \\
		\midrule
		\multirow{7}{*}{TinyImageNet} 
		         & \multirow{7}{*}{WRN34X10} 
		         & AT$^\dagger$
		         & 30.65 & 6.81 \\
		         & & LBGAT$^\dagger$
		         & 36.50 & 14.00 \\
		         & & ALP \citep{kannan2018adversarial}$^\dagger$
		         & 30.51 & 8.01 \\
		         & & TRADES$^\dagger$
		         & 38.51 & 13.48 \\
		         & & LBGAT+ALP$^\dagger$
		         & 33.67 & 14.55 \\
		         & & LBGAT+TRADES$^\dagger$
		         & 39.26 & 16.42 \\
		         & & \textbf{DIO}+AT
		         & $\mathbf{46.02\pm0.17}$ 
		         & $\mathbf{22.81\pm0.15}$ \\
		\bottomrule
	\end{tabular}
\end{table*}

\section{Supplementary empirical results on adversarial robustness}
\subsection{Results on CIFAR100 and TinyImageNet}

Aside from the robust performance on CIFAR10 
in the manuscript, we provide more results on CIFAR100 and TinyImageNet in terms of both clean and robust accuracy in Table \ref{tab:robustness-comparison-3} for a comprehensive comparison with more defenses \citep{rice2020overfitting,sitawarin2021sat,chen2022efficient,kannan2018adversarial}.
In Table \ref{tab:robustness-comparison-3}, for CIFAR10 and CIFAR100, the adversarial robustness of various defenses is evaluated against the ensemble attack of AA, while the robust accuracy on TinyImageNet corresponds to the white-box PGD-20 attack.
As shown, in the case of more complex data, when considering widely-used data augmentation techniques (AT and TRADES), the adversarial robustness of the neural networks can get enhanced by the model augmentation from  DIO.
For instance, on TinyImageNet, the proposed DIO, which is adversarially trained under PGD-10 attack, achieves over 20\% robust accuracy against the PGD-20 attack and maintains 46.02\% clean accuracy,  outperforming other adversarial defenses distinctively.

\begin{table}[!ht]
	\centering
	\caption{Comparison results of the clean and robust accuracy (\%) of defenses trained with and without the DDPM-generated data on CIFAR10 and CIFAR100. The DDPM data augmentation brings substantial adversarial robustness gains, on the basis of which DIO shows even more adversarial robustness improvements.}\label{tab:ddpm}
 	\resizebox{\textwidth}{!}{
	\begin{tabular}{c|c|c|cccc}
		\toprule
        data set & defense & additional data & clean & PGD-20 & PGD-100 & AA \\
		\midrule
		\multirow{3}{*}{CIFAR10} & TRADES & / & 80.21 & 51.92 & 51.83 & 48.61\\
		& TRADES & DDPM & 80.74 & 56.40 & 56.29 & 52.86\\
		& \textbf{DIO}+TRADES & DDPM & $\bf 81.64\pm0.06$ & $\bf 57.00\pm0.04$ & $\bf 56.89\pm0.04$ & $\bf 53.83\pm0.07$ \\
        \midrule
		\multirow{3}{*}{CIFAR100} & TRADES & / & 54.53 & 27.85 & 27.79 & 23.61\\
		& TRADES & DDPM & 53.55 & 30.00 & 29.98 & 25.68\\
		& \textbf{DIO}+TRADES & DDPM & $\bf 56.48\pm0.05$ & $\bf 31.07\pm0.05$ & $\bf 31.03\pm0.05$ & $\bf 28.20\pm0.06$ \\
		\bottomrule
	\end{tabular}
 	}
\end{table}

\subsection{Results on DDPM-generated data}

In this subsection, we explore the robustness of DIO equipped with augmented data generated by the prevailing Denoising Diffusion Probabilistic Models (DDPMs, \citep{ho2020denoising}).
As a new type of generative models, DDPM has shown powerful abilities in generating even more verisimilar images than traditional GANs \citep{dhariwal2021diffusion}, and recently many amazing artworks have been created via DDPM-based algorithms in AI painting, such as stable diffusion \citep{rombach2022high}.
In regards to adversarial robustness, \citet{gowal2021improving} involved DDPM-generated data into adversarial training as a strong data augmentation technique and achieved new SOTA results on multiple data sets, which ranks the first \citep{rebuffi2021fixing} on the RobustBench\footnote{\href{https://robustbench.github.io/}{https://robustbench.github.io/}\label{fn:robustbench}}.

DIO is a model-augmentation adversarial defense method and is compatible with various data augmentation techniques. 
Preceding experiments have shown that when the standard adversarial training is executed on those training sets of CIFAR10, CIFAR100 and TinyImageNet, the model-augmentation DIO manifests stronger adversarial robustness than a regular DNN.
Apart from this, in this section, we further show that when generated data is included into the training set for adversarial training, DIO could even reach more adversarial robustness gains.

\noindent\textbf{Settings}\quad Following the basic settings adpoted by the top-1 method \citep{rebuffi2021fixing,gowal2021improving} in RobustBench\textsuperscript{\ref{fn:robustbench}}, the additional training data is generated via 2 DDPMs that are solely trained on the original training sets of CIFAR10 and CIFAR100. 
1M images are sampled from the trained DDPMs and included into the training sets of CIFAR10 and CIFAR100 respectively.
TRADES \citep{zhang2019theoretically} is used as the adversarial training technique with an SGD optimizer with Nesterov momentum 0.9 and a global weight decay of $5e^{-4}$.
We train 400 epochs for each experiment and use a cosine learning rate schedule with an initial value of 0.1.
The selected model architecture is WRN28X10 \citep{zagoruyko2016wide} with Swish/SiLU activations \citep{hendrycks2016gaussian}.
For DIO, we set $\alpha=0.1,\beta=0.1,\tau=0.1,L=10$ and set the training batch size to 400 with a ratio of original-to-generated data of 0.3.
All the experiments are executed on a single A100 GPU with 40 GiB memory.

Notice that our target is neither to reproduce the explicit results reported in \citep{rebuffi2021fixing,gowal2021improving} nor to create new SOTAs, as we are unable to obtain as many GPUs as the required 32 Google TPUv3 cores or as large a batch size up to 1024.
More importantly, this experiment aims to demonstrating that DIO could further benefit adversarial robustness more even with the help of additional DDPM-generated training data.

\noindent\textbf{Results}\quad Tab.\ref{tab:ddpm} illustrates the comparison results among different defenses and data augmentation techniques on CIFAR10 and CIFAR100.
To be specific, we compare (i) the performance of TRADES trained with and without DDPM-generated data, and (ii) the performance of DIO trained with DDPM-generated data.
As indicated in Tab.\ref{tab:ddpm}, in the case of a standard DNN, the additional DDPM-generated data benefits a lot the adversarial robustness, which improves the PGD-100 accuracy on CIFAR10 from 51.83\% to 56.29\%.
Besides, once the model augmentation DIO is introduced, both the robust accuracy and the clean accuracy gets even more boosted, illustrating the superiority of such a model augmentation in promoting adversarial robustness under the case of the DDPM data augmentation.

\subsection{Results against adaptive attacks}

To further evaluate the adversarial robustness,  we design and execute 2 different adaptive attacks. DIO has multiple mutually-orthogonal heads and randomly selects one head in the forward process. 
With considerations to the network topology and forward strategy of DIO, one can naturally raise two questions: 
\begin{itemize}
    \item[(i)] Whether the randomness gives a false sense of security, as the gradient masking discussed in \citep{athalye2018obfuscated}?
    \item[(ii)] Does each individual network {\it w.r.t.} each head in DIO still hold strong adversarial robustness?
\end{itemize}

To answer such 2 questions, we design 2 adaptive attacks as follows, where the randomness in DIO is eliminated in 2 different ways.

\noindent \textbf{Adaptive-Attack-1 (Adapt-A1):} The solution to eliminate the randomness in \citep{athalye2018obfuscated} is to compute the expectation over multiple instantiations of randomness (see Sec.5.3 in \citep{athalye2018obfuscated}). 
Therefore, during attack, at each iteration of gradient descent, we use the average loss of all the heads to compute the gradient direction, so that the iteration of PGD is reformulated as:
\begin{equation}
    \label{eq:PGD-adapt-1}
    \mathbf{\hat{x}}^{(t+1)} = P_\epsilon\left\{\mathbf{\hat{x}}^{(t)} + \gamma\cdot\mathrm{sign}\left(\nabla_{\mathbf{x}}\frac{1}{L}\sum_{i=1}^{L}\mathcal{L}(h^i(g(\mathbf{\hat{x}}^{(t)})),y)\right)\right\}.
\end{equation}
Considering that DIO is {\it NOT an ensemble defense} in the sense of randomly selecting one head during inference, Adapt-A1 is devised as an {\it ensemble attack} of utilizing all heads to perform attacks, so as to get rid of the influence of the gradient masking in generating adversarial examples.
After the adversarial examples are generated by Adapt-A1, these perturbed images are then recognized by the $L$ heads in DIO, respectively, resulting in $L$ robust accuracy.
We report all the $L$ robust accuracy in the following Fig.\ref{fig:exp-adaptive-attack} and Fig.\ref{fig:exp-adaptive-attack-appendix}.

\noindent \textbf{Adaptive-Attack-2 (Adapt-A2):} Aside from Adapt-1 involving all heads to generate the adversarial examples, we alternate to evaluate the adversarial robustness of each individual path.
That is, each of the $L$ heads $h^i(\cdot),i=1,\cdots,L$ could collaborate with the shared backbone $g(\cdot)$ to constitute a standard neural network, {\it i.e.}, each individual path indicates a standard DNN such that $f^i(\cdot)\triangleq h^i(g(\cdot)),i=1,\cdots,L$, then, the PGD attack is executed on such a standard network $f^i(\cdot)$:
\begin{equation}
    \label{eq:PGD-adapt-2}
    \mathbf{\hat{x}}^{(t+1)} = P_\epsilon\left\{\mathbf{\hat{x}}^{(t)} + \gamma\cdot\mathrm{sign}\left(\nabla_{\mathbf{x}}\mathcal{L}({\color{red}{f^i}}(\mathbf{\hat{x}}^{(t)}),y)\right)\right\}.
\end{equation}
Accordingly, Adapt-A2 also produces $L$ robust accuracy in correspondence to the $L$ paths in DIO.
These $L$ robust accuracy is reported in Fig.\ref{fig:exp-adaptive-attack} and Fig.\ref{fig:exp-adaptive-attack-appendix}.

Fig.\ref{fig:exp-adaptive-attack} and Fig.\ref{fig:exp-adaptive-attack-appendix} give the results, where two strong white-box attacks, PGD-20 and PGD-100, are involved into the adaptive attacks to evaluate the robustness of DIO.
We select two DIO models of WRN34X10 trained with different data augmentation techniques, {\it i.e.}, ``DIO+AT'' with $L=10$ and ``DIO+GAIRAT'' with $L=40$.
The $L$ robust accuracy in correspondence to the $L$ paths is reported, indexed by the x-axes in Fig.\ref{fig:exp-adaptive-attack} and Fig.\ref{fig:exp-adaptive-attack-appendix}.

\begin{figure*}[!ht]
\centering
\includegraphics[width=1.0\linewidth]{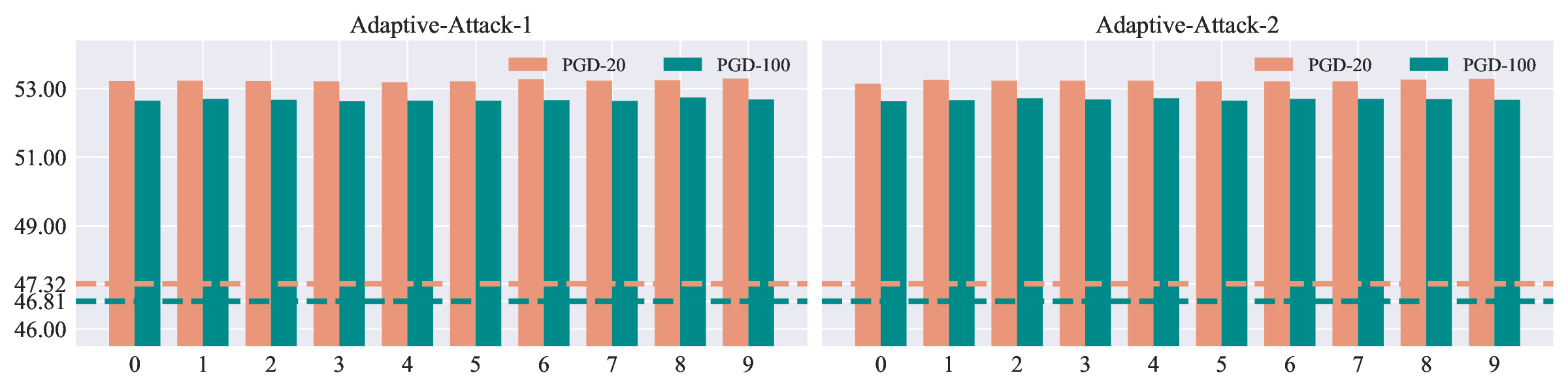}
\caption{Results of two adaptive attacks: Each bar in the charts indicates the robust accuracy (\%) of each path in DIO (\textbf{``DIO+AT''} of \textbf{WRN34X10}) against the white-box PGD-20 (the salmon bars) and PGD-100 (the cyan bars) attacks.
The left chart illustrates the results of Adapt-A1, which averages the losses of all  paths to compute the gradients.
The right chart illustrates the results of Adapt-A2, which is executed directly on the each individual network.
The salmon and cyan dashed lines indicate the baseline results of the standard DNN (\textbf{``AT''} of \textbf{WRN34X10}) against PGD-20 and PGD-100, respectively.}
\label{fig:exp-adaptive-attack}
\end{figure*}

\begin{figure*}[!h]
\centering
\includegraphics[width=1.0\linewidth]{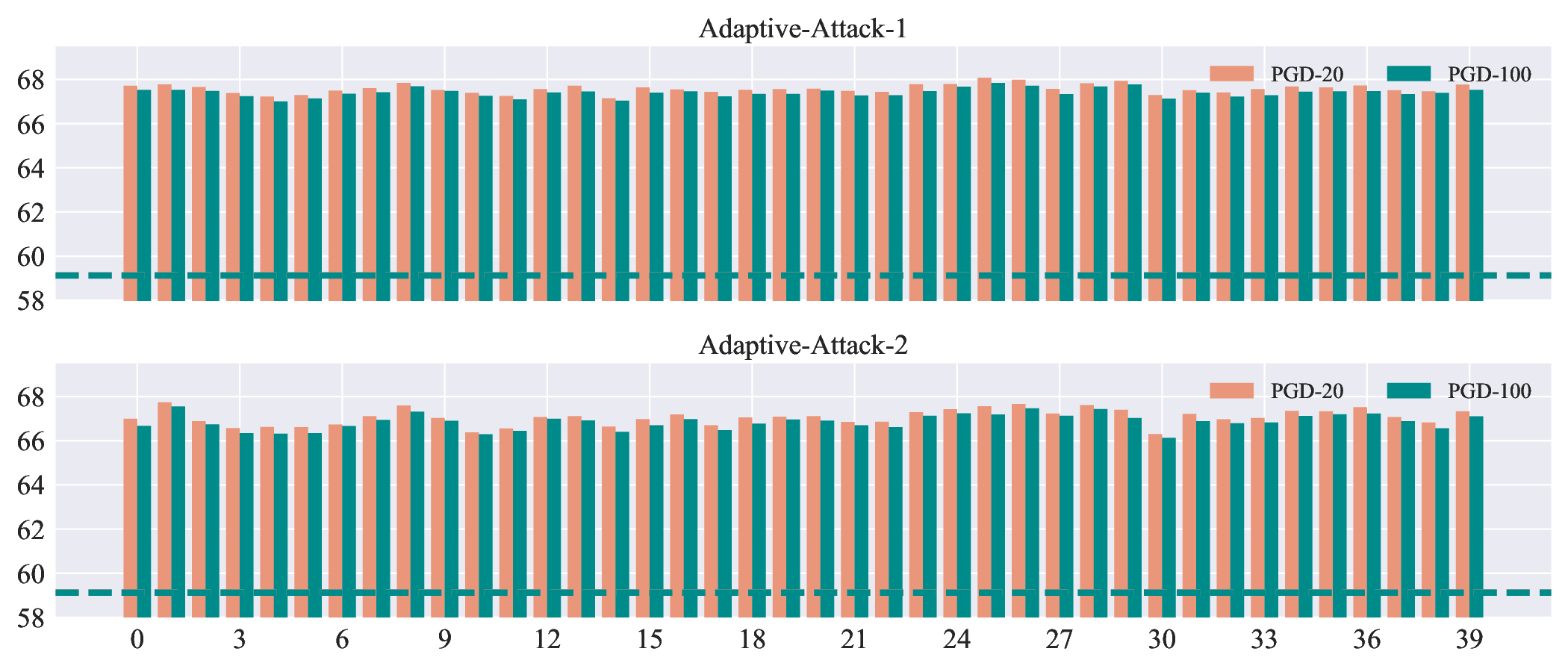}
\caption{Results of two adaptive attacks: Each bar in the charts indicates the robust accuracy (\%) of each path in DIO (\textbf{``DIO+GAIRAT''} of \textbf{WRN34X10}) against the white-box PGD-20 (the salmon bars) and PGD-100 (the cyan bars) attacks.
The top chart illustrates the results of Adapt-A1, which averages the losses of all the paths to compute the gradients.
The bottom chart illustrates the results of Adapt-A2, which is executed directly on the each individual network.
The two nearly-overlapping dashed salmon and cyan lines indicate the baseline results of the standard neural network (\textbf{``GAIRAT''} of \textbf{WRN34X10}) against PGD-20 and PGD-100, {\it i.e.}, 59.13 and 59.12, respectively.}
\label{fig:exp-adaptive-attack-appendix}
\end{figure*}

Under the Adapt-A1 and Adapt-A2, the influence of gradient masking is completely eliminated in different ways, while each individual path in DIO still manifests superior robustness performance over the baseline models, shown in Fig.\ref{fig:exp-adaptive-attack} and Fig.\ref{fig:exp-adaptive-attack-appendix}.
Such results against adaptive attacks can answer the two questions raised before.
The strong adversarial robustness of DIO is not caused by the gradient masking \citep{athalye2018obfuscated}.
All the individual networks in DIO actually gets \textit{consistently and substantially} benefited from the proposed model augmentation.

Besides, the enhanced robustness is attributed {\it the most} to the learned robust features by the proposed multi-head structure with orthogonality and distance constraints.
The Adapt-A2, which evaluates the robustness of each individual path with other heads pruned, is devised exactly to illustrate the effectiveness of the learned robust features, as these paths hold the same backbone to extract features and only differ in the last layer (head).

\subsection{List of the hyper-parameters}

As shown in Alg.\ref{alg}, the training of DIO requires 4 hyper-parameters: the number of heads $L$, the upper bound of the distance $\tau$ and the coefficients $\alpha$ and $\beta$ to control the penalties of the losses $\mathcal{L}_o$ and $\mathcal{L}_d$, respectively. 
In Table \ref{tab:hyper-parameters}, we list the detailed values of these hyper-parameters  corresponding to the various DIO models appearing in the Table 1 and Table 2 in the manuscript and in Table \ref{tab:robustness-comparison-3} in this supplementary material for reproducibility.

\begin{table}[!h]
	\centering
	\caption{Hyper-parameters $\alpha$, $\beta$, $\tau$ and $L$ of various DIO models appearing in the adversarial robustness evaluation.}\label{tab:hyper-parameters}
 	\scalebox{0.9}{
	\begin{tabular}{c|cc|cccc}
		\toprule
        defense & data set & model & $\tau$ & $L$ & $\alpha$ & $\beta$ \\
		\midrule
        \multirow{2}{*}{DIO} 
        & CIFAR10  
        & \multirow{2}{*}{PRN18} 
        & $0.2$ & $10$ 
        & \multirow{2}{*}{$0.1$} 
        & \multirow{2}{*}{$0.1$} \\
          & CIFAR100 & & $0.01$ & $40$ & & \\
        \midrule
        \multirow{5}{*}{DIO+AT}
        & CIFAR10 & PRN18 & $0.2$ & $10$ 
        & \multirow{16}{*}{$0.1$} 
        & \multirow{16}{*}{$0.1$} \\
        & CIFAR10 & WRN34X10 & $0.5$ & $10$ & &  \\
        & CIFAR100 & WRN34X10 & $0.1$ & $10$ & & \\
        & CIFAR100 & WRN34X20 & $0.2$ & $10$ & & \\
        & TinyImageNet & WRN34X10 & $0.1$ & $10$ & & \\
        \cmidrule{2-5}
        \multirow{4}{*}{DIO+TRADES}
        & CIFAR10 & PRN18 & $0.1$ & $10$ & & \\
        & CIFAR10 & WRN34X10 & $0.1$ & $10$ & & \\
        & CIFAR10 & WRN34X20 & $0.1$ & $10$ & & \\
        & CIFAR100 & WRN34X10 & $0.2$ & $10$ & & \\
        \cmidrule{2-5}
        \multirow{2}{*}{DIO+AWP}
        & \multirow{6}{*}{CIFAR10} & PRN18 
        & $0.001$ & $20$ & & \\
         & & WRN34X10  
        & $0.01$ & $10$ & & \\
        \multirow{2}{*}{DIO+LBGAT}
        & & PRN18 & $0.1$ & $10$ & & \\
        & & WRN34X10 & $0.1$ & $10$ & & \\
        \multirow{2}{*}{DIO+GAIRAT}
        & & PRN18 & $0.02$ & $40$ & & \\
        & & WRN34X10 & $0.02$ & $40$ & & \\
		\bottomrule
	\end{tabular}
 	}
\end{table}

\section{Supplementary empirical analysis}

\subsection{Comparisons on model capacities}

It is widely acknowledged that models with larger capacities, {\it i.e.}, more trainable parameters, tend to produce stronger adversarial robustness.
Meanwhile, the multiple heads in the proposed DIO increase the model capacity, which should be taken into consideration in robustness evaluations.
Accordingly, we report the number of parameters and heads for those adversarially-trained DIO models in this paper and compare with the baseline standard models, shown in Tab.\ref{tab:model-capacity-comparison} below.

\begin{table*}[!htbp]
	\centering
	\caption{Comparisons on the model capacity between the baseline standard models and the DIO multi-head models on different data sets and network structures.}\label{tab:model-capacity-comparison}
 	\resizebox{\textwidth}{!}{
	\begin{tabular}{c|c|c|c|c}
		\toprule
		data set with $K$ classes & model & baseline ($L=1$) & DIO with $L$ heads & increasement \\
		\midrule
		\multirow{6}{*}{CIFAR10 ($K=10$)} & \multirow{3}{*}{PRN18} & \multirow{3}{*}{11,172,170} & 11,218,240 ($L=10$) & 0.4124\%\\
		& & & 11,269,440 ($L=20$) & 0.8706\%\\
		& & & 11,371,840 ($L=40$) & 1.7872\%\\
		\cmidrule{2-5}
		& \multirow{2}{*}{WRN34X10} & \multirow{2}{*}{46,160,474} & 46,218,064 ($L=10$) & 0.1248\%\\
		& & & 46,410,064 ($L=40$) & 0.5407\%\\
		\cmidrule{2-5}
		& WRN34X20 & 184,531,674 & 184,646,864 ($L=10$) & 0.0624\%\\
		\midrule
		\multirow{2}{*}{CIFAR100 ($K=100$)} & WRN34X10 & 46,218,164 & 46,794,064 ($L=10$) & 1.2460\% \\
		\cmidrule{2-5}
		& WRN34X20 & 184,646,964 & 185,798,864 ($L=10$) & 0.6238\% \\
		\midrule
		TinyImageNet ($K=200)$ & WRN34X10 & 46,282,264 & 47,434,064 ($L=10$) & 2.4886\%\\
		\bottomrule
	\end{tabular}}
\end{table*}

As shown in Tab.\ref{tab:model-capacity-comparison}, the number of the increased extra parameters brought by the multi-head structure of DIO is very minor (generally less than 2 percentages) compared with the baseline standard model, since the parameter number of the last linear layer (head) is much less than that of the preceding convolution layers (backbone).
Such minor capacity increasement actually cannot yield significant gains on the model performance.
Therefore, we can rule out the effect in robustness improvements of such a factor of the increased model capacity of DIO.

\subsection{Convergence analysis}
\label{sec:exp-convergence}
\begin{figure*}[!h]
\centering
\subfloat[Cross entropy]{\includegraphics[width=0.45\linewidth]{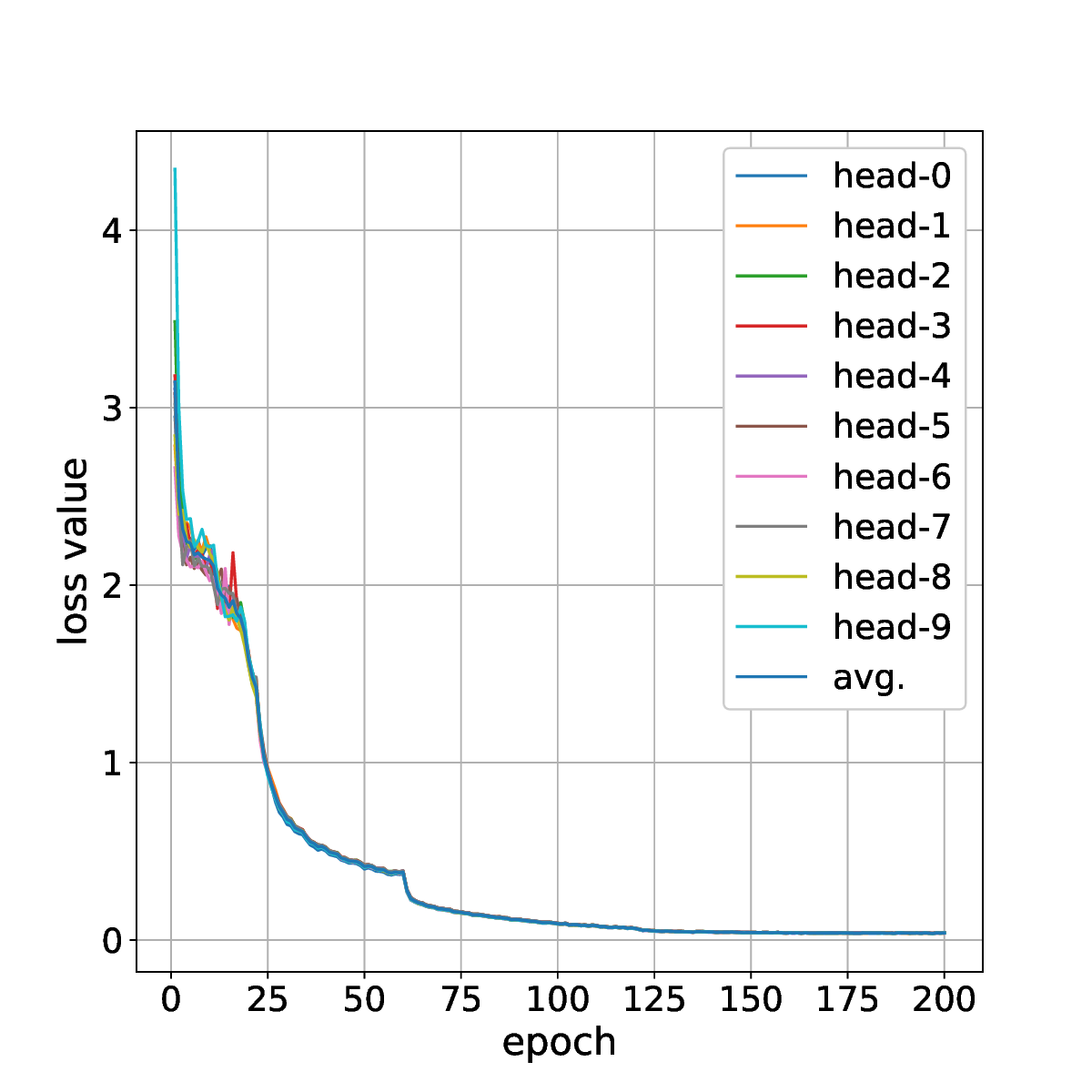}%
\label{fig:ce-loss}}
\hfil
\subfloat[Orthogonality constraint $\mathcal{L}_{o}$]{\includegraphics[width=0.45\linewidth]{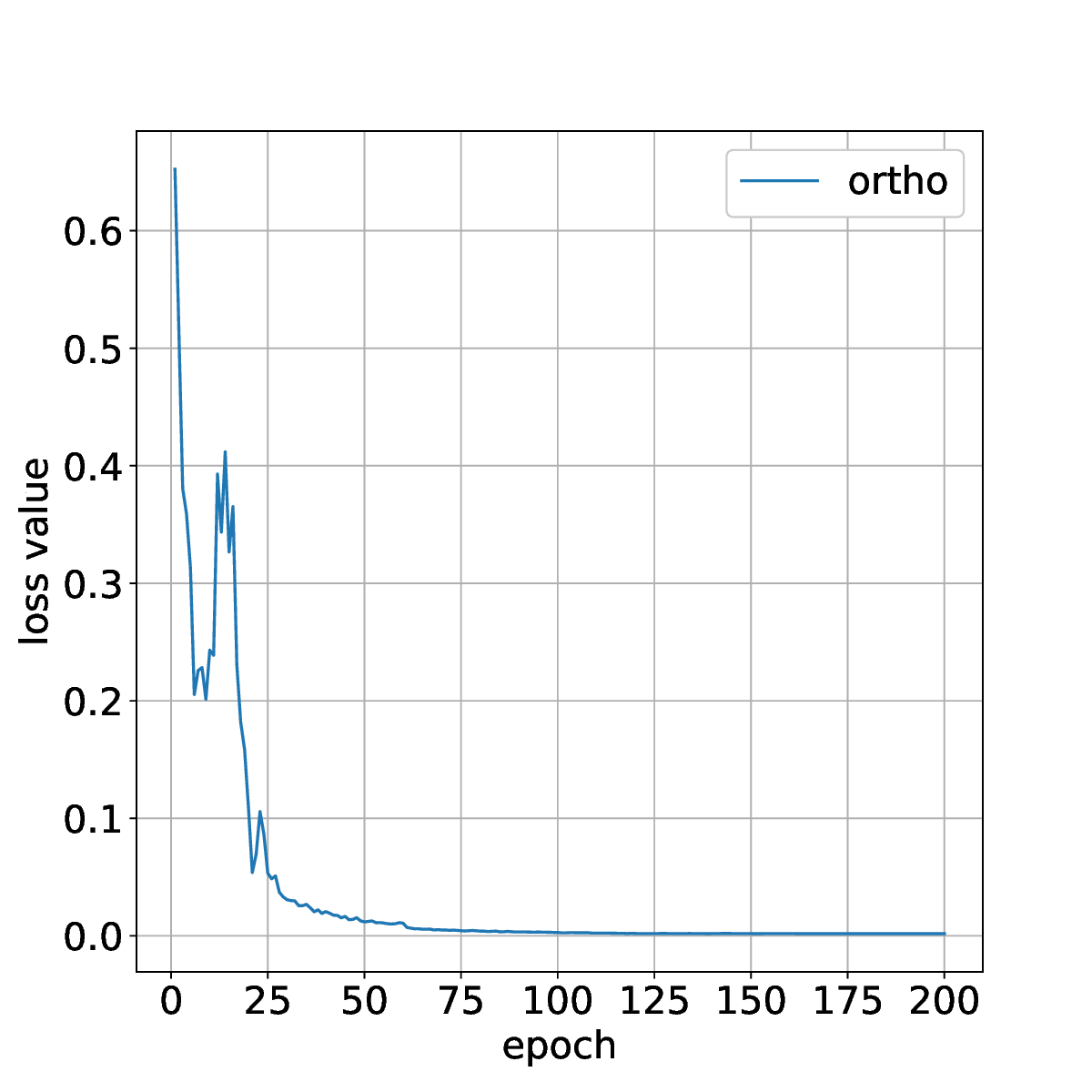}%
\label{fig:ortho-loss}}
\vfil
\subfloat[Distance constraint $\mathcal{L}_{d}$]{\includegraphics[width=0.45\linewidth]{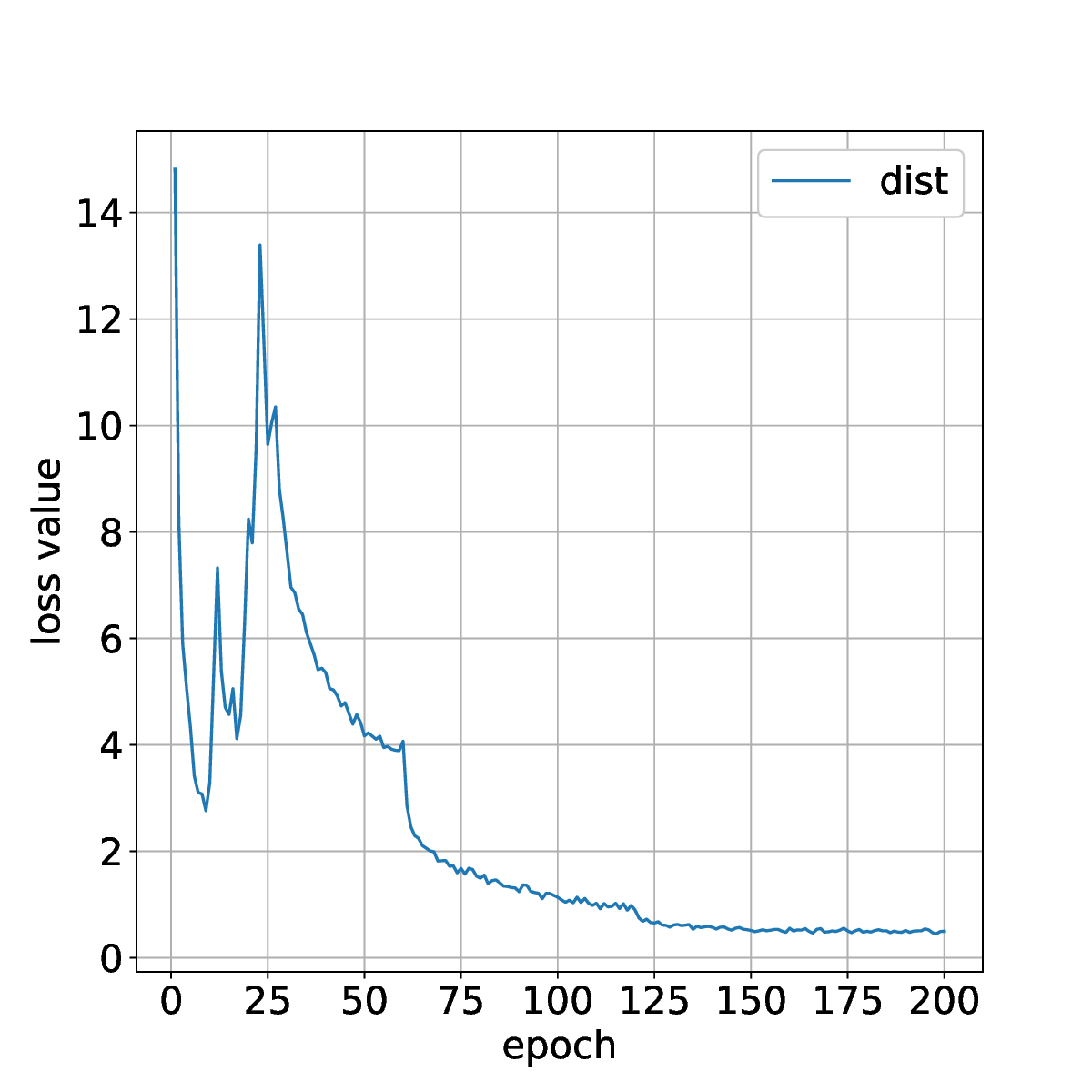}%
\label{fig:dist-loss}}
\hfil
\subfloat[ROC curves]{\includegraphics[width=0.45\linewidth]{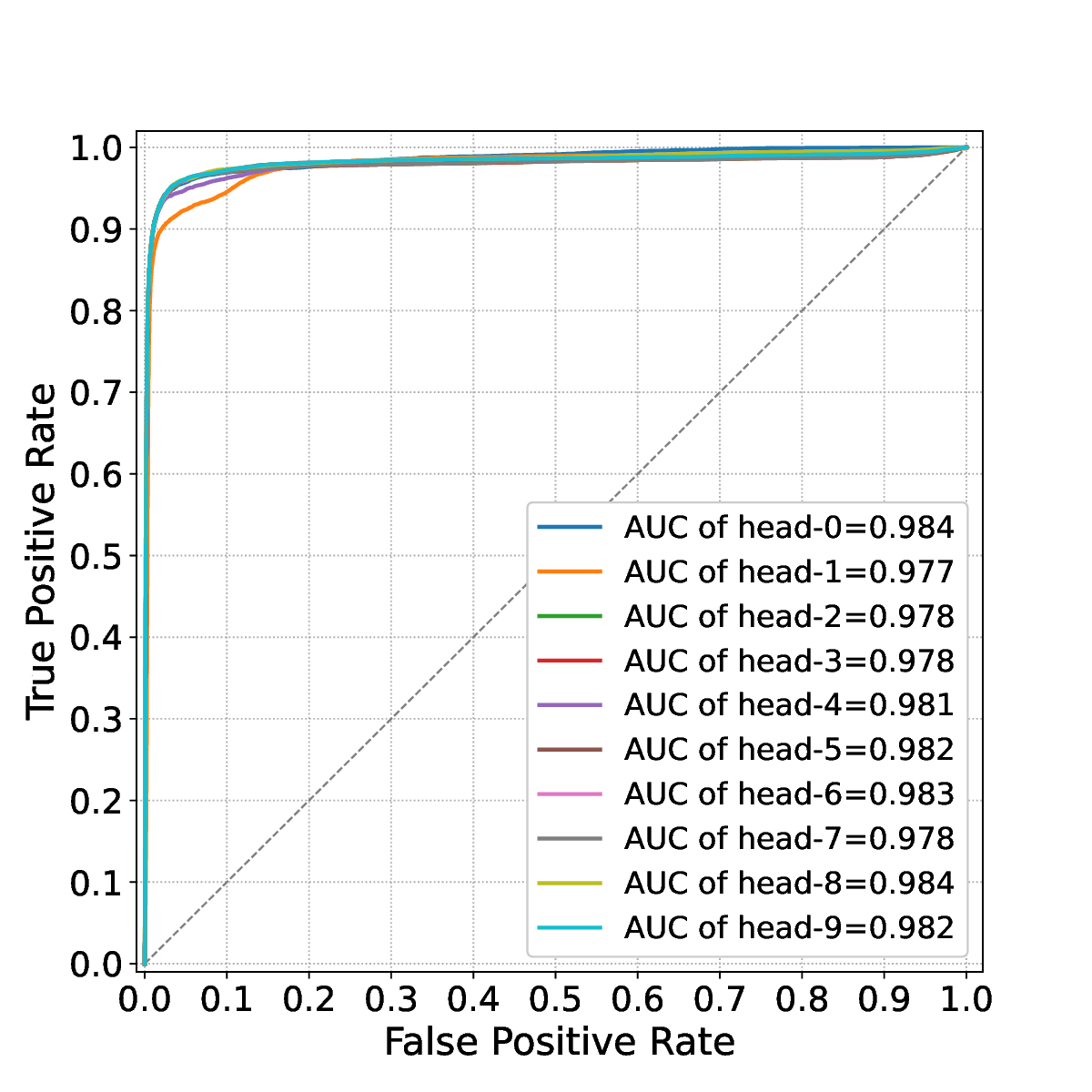}%
\label{fig:roc-curves}}
\caption{Loss curves during the training of DIO and the ROC curves of a well-trained DIO model.}
\label{fig:loss-curves}
\end{figure*}

In this subsection, a series of experiments are conducted  to investigate the properties of the  individual networks in DIO {\it w.r.t.} each head in both the training and inference phases.
Fig.\ref{fig:loss-curves} shows the loss curves of DIO during the training phase on CIFAR10, including the cross entropy loss $\mathcal{L}_{ce}$ {\it w.r.t.} each head (Fig.\ref{fig:ce-loss}), the orthogonality constraint $\mathcal{L}_{o}$ (Fig.\ref{fig:ortho-loss}) and the distance constraint $\mathcal{L}_{d}$ (Fig.\ref{fig:dist-loss}).
Besides, in the test phase, the  ROC curves and AUC values of each individual network on the test set of CIFAR10 are plotted in Fig.\ref{fig:roc-curves}. 

In Fig.\ref{fig:ce-loss}, at the beginning of the training phase, there exists a discrepancy among the cross entropy loss curves of different heads.
As the training develops, the multiple cross entropy loss curves gradually converge to nearly the same one.
In Fig.\ref{fig:ortho-loss} and Fig.\ref{fig:dist-loss}, the orthogonality constraint $\mathcal{L}_o$ and distance constraint $\mathcal{L}_d$ both can reduce to very small values, indicating that the training of such a multi-head structure with such 3 losses can successfully converge.
In Fig.\ref{fig:roc-curves}, to draw the ROC curve of each individual network $h^i(g(\cdot))$ on the 10-classification test set, 10 groups of FPR and TPR values are first determined on each class of the test data via the one-vs-rest strategy.
Then the ROC curve is obtained by averaging these 10 groups of FPR and TPR values.
It is obvious  that each individual network in DIO manifests nearly the same ROC curve and very close AUC value.
The fact that the loss curves in training and the ROC curves in inference further support  that the individual networks in DIO show quite approximative generalization performance.

Combining the generalization performance and the robustness performance together, we have the following conclusion:
\textit{The augmented diversity via orthogonality does not hurt the generalization performance of each individual network, and yet simultaneously benefits  each individual network with a substantial improvement on  robustness}.

\subsection{Sensitivity analysis on $\tau$}

As the Eq.(9) and Eq.(10) in the manuscript indicate, the margin $D(\mathbf{\bar{z}},h)$ measures the minimal distance required to change the current correct prediction, and the loss $\mathcal{L}_d$ measures the approximation between such a minimal distance and the hyper-parameter $\tau$.
Therefore, $\tau$ actually indicates the upper bound of the minimal distance to some extent,  naturally leading to an inference, {\it i.e.}, larger $\tau$ might bring better robustness. In this regard, corresponding numerical experiments are performed, in which  the performance variations of different DIO models {\it w.r.t.} different $\tau$ against FGSM attack are evaluated. Two vanilla-trained models, VGG11 on CIFAR10 and WRN28X5 on CIFAR100, are considered.
The considered sets of $\tau$ are $\{0.5,1.0,2.0,3.0,4.0\}$ and $\{0.1,0.2,0.5,1.0\}$, respectively.

\begin{figure}[!h]
\centering
\subfloat[VGG11 trained on CIFAR10]{\includegraphics[width=0.45\linewidth]{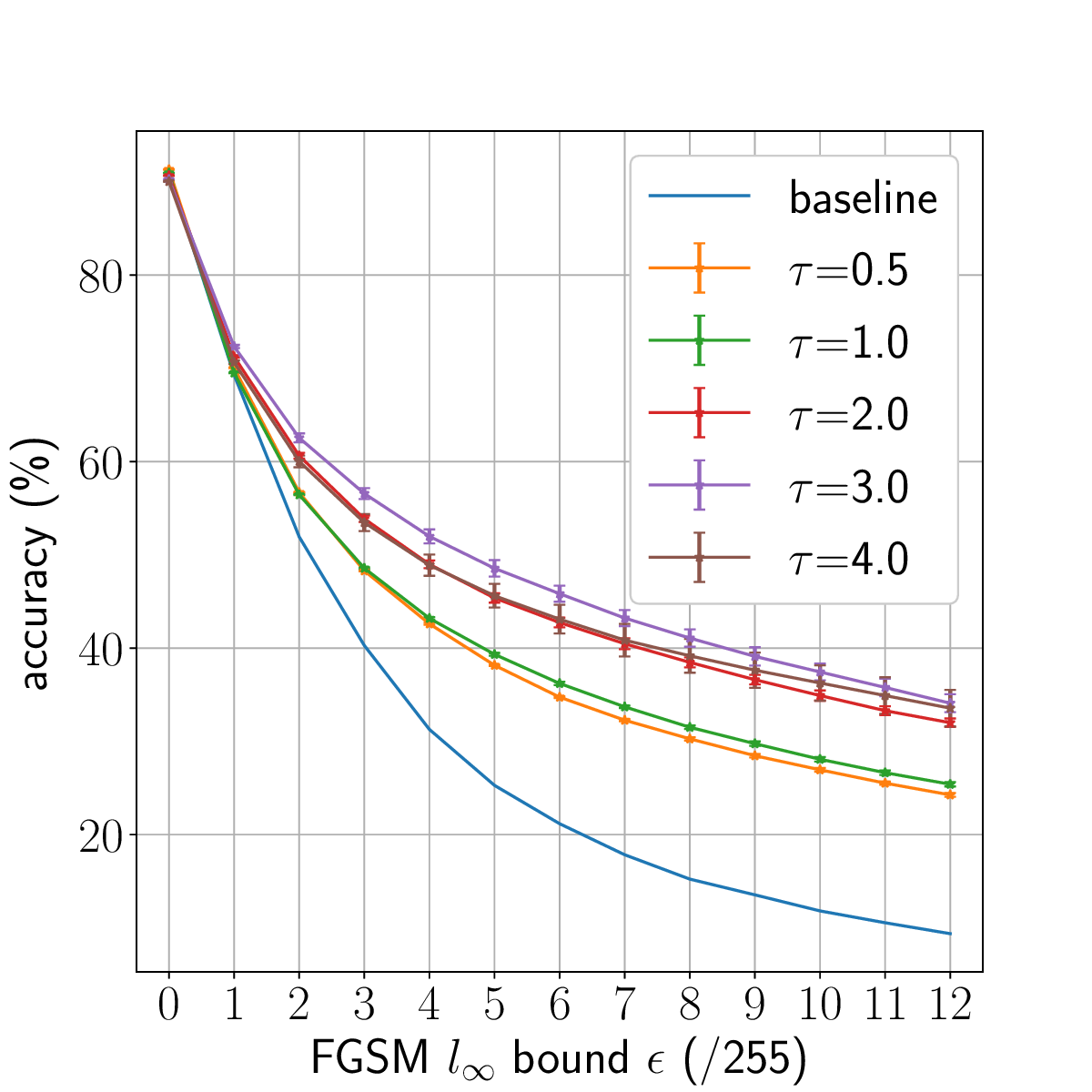}%
\label{fig:tau-vgg11}}
\hfil
\subfloat[WRN28X5 trained on CIFAR100]{\includegraphics[width=0.45\linewidth]{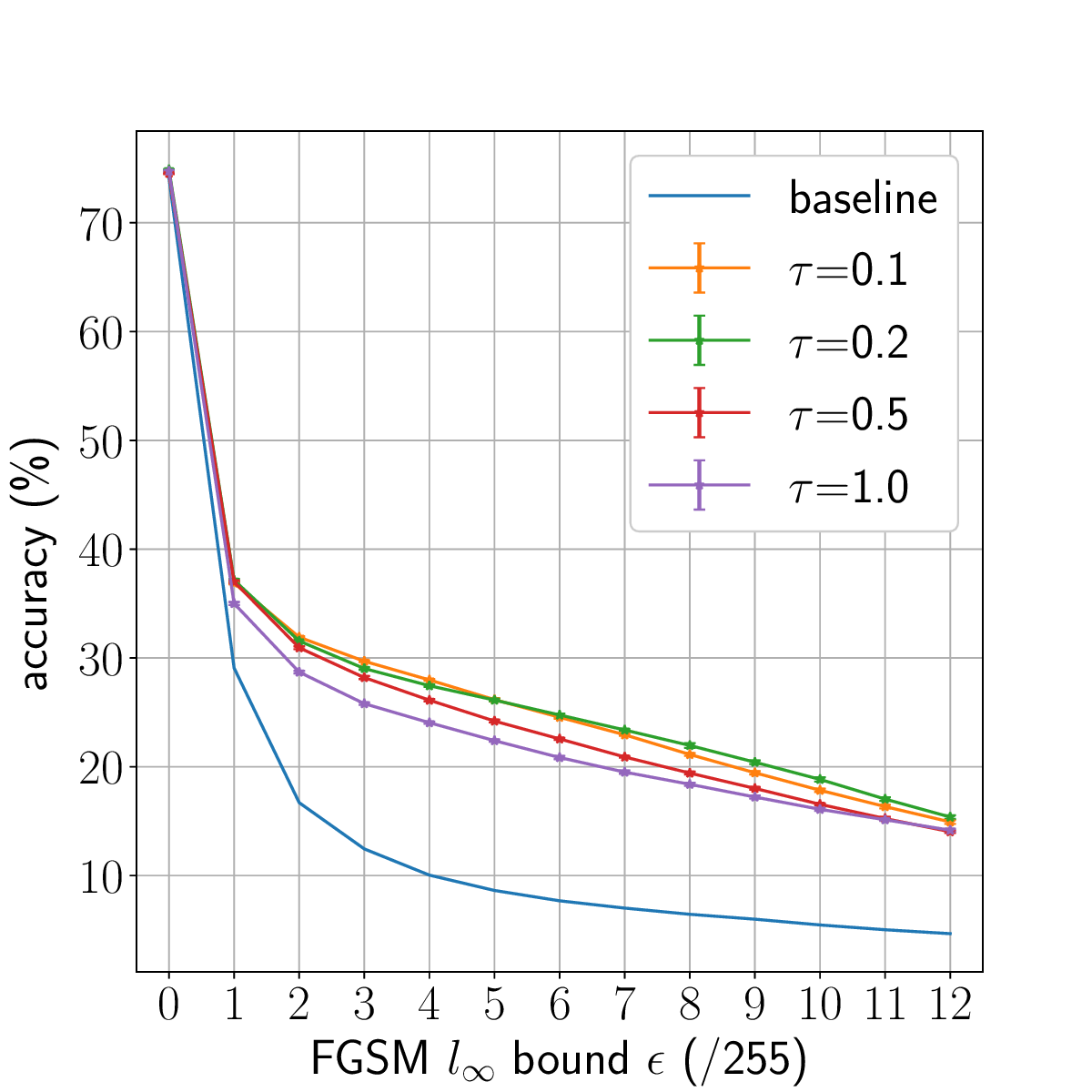}%
\label{fig:tau-wrn28x5}}
\caption{Adversarial robustness of DIO {\it w.r.t.} different values of $\tau$.}
\label{fig:tau-effect}
\end{figure}

Fig.\ref{fig:tau-effect} shows the  results concerning the effect of $\tau$ against the FGSM attack.
For DIO of VGG11, the best model robustness is achieved at $\tau=3.0$.
For DIO of WRN28X5, the best model robustness is achieved at $\tau=0.2$.
Accordingly, within a certain range, as the $\tau$ increases, the robustness of DIO gets improved, which is in accordance with our inference, but the improvements seem limited. Thus,
a very large value of $\tau$ can have a side effect on  the robustness, and empirical validations are suggested to make a mild determination of of $\tau$ under varied settings.

\end{appendices}
\end{document}